\documentclass[10pt,twocolumn,letterpaper]{article}

\usepackage{wacv}              

\definecolor{wacvblue}{rgb}{0.21,0.49,0.74}
\usepackage[pagebackref,breaklinks,colorlinks,allcolors=wacvblue]{hyperref}
\usepackage{graphicx}
\usepackage{wrapfig}
\usepackage{float}
\usepackage{placeins}
\usepackage{tabularx}
\usepackage{subcaption}
\usepackage{caption}
\usepackage{colortbl}
\usepackage{xcolor}
\usepackage{array} 
\usepackage{amssymb} 

\usepackage{url}            
\usepackage{booktabs}       
\usepackage{amsfonts}       
\usepackage{nicefrac}       
\usepackage{microtype}      
\usepackage{makecell}       
\usepackage{multicol} 
\usepackage{multirow}
\usepackage{makecell}
\usepackage{xspace}
\usepackage{multirow}
\usepackage{adjustbox}
\usepackage{colortbl}
\usepackage{duckuments}
\usepackage{lipsum}
\usepackage{enumitem}
\usepackage{kotex}
\usepackage{tikz}
\usepackage{amsmath}
\usepackage{subcaption}
\usepackage{graphicx}
\usepackage{tabularx}
\newcolumntype{Y}{>{\centering\arraybackslash}X}
\usepackage{rotating}
\usepackage{color}
\usepackage[utf8]{inputenc}

\usepackage{amsthm}

\def\wacvPaperID{1902} 
\def\confName{WACV}
\def\confYear{2026}

\title{Towards Reliable Test-Time Adaptation: \\ Style Invariance as a Correctness Likelihood}

\author{Gilhyun Nam$^\textrm{*1}$ \quad
Taewon Kim$^\textrm{*1}$ \quad
Joonhyun Jeong$^\textrm{1,2}$ \quad
Eunho Yang$^\textrm{\textdagger 1,3}$\\
$^\textrm{1}$KAIST\quad
$^\textrm{2}$NAVER Cloud \quad
$^\textrm{3}$AITRICS \\
{\tt\small \{gilhyun.nam, maxkim139, joonhyun.jeong, eunhoy\}@kaist.ac.kr}
}

\begin{document}
\maketitle
\begin{abstract}

Test-time adaptation (TTA) enables efficient adaptation of deployed models, yet it often leads to poorly calibrated predictive uncertainty—a critical issue in high-stakes domains such as autonomous driving, finance, and healthcare. Existing calibration methods typically assume fixed models or static distributions, which leads to degraded performance under real-world, dynamic test conditions. To address these challenges, we introduce \textbf{S}tyle \textbf{I}nvariance as a \textbf{C}orrectness \textbf{L}ikelihood (SICL), a framework that leverages style-invariance for robust uncertainty estimation. SICL estimates instance-wise correctness likelihood by measuring prediction consistency across style-altered variants, requiring only the model’s forward pass. This makes it a plug-and-play, backpropagation-free calibration module compatible with any TTA method. Comprehensive evaluations across four baselines, five TTA methods, and two realistic scenarios with three model architectures demonstrate that SICL reduces calibration error by an average of 13\%p compared to conventional calibration approaches.
\end{abstract}    
\setcounter{footnote}{1}
\footnotetext{\textsuperscript{*}Equal Contribution. \textsuperscript{\textdagger}Corresponding Author.}
\section{Introduction}\label{sec:intro}
Deep learning has revolutionized numerous domains, yet the real-world deployment of machine learning models continues to face a critical challenge: performance degradation due to distributional shifts. These shifts occur when deployment conditions deviate from the training environment, including variations in environmental conditions, encounters with unseen objects, or the presence of unexpected noise.

\begin{figure}[t!]
    \centering
    \includegraphics[width=\columnwidth]{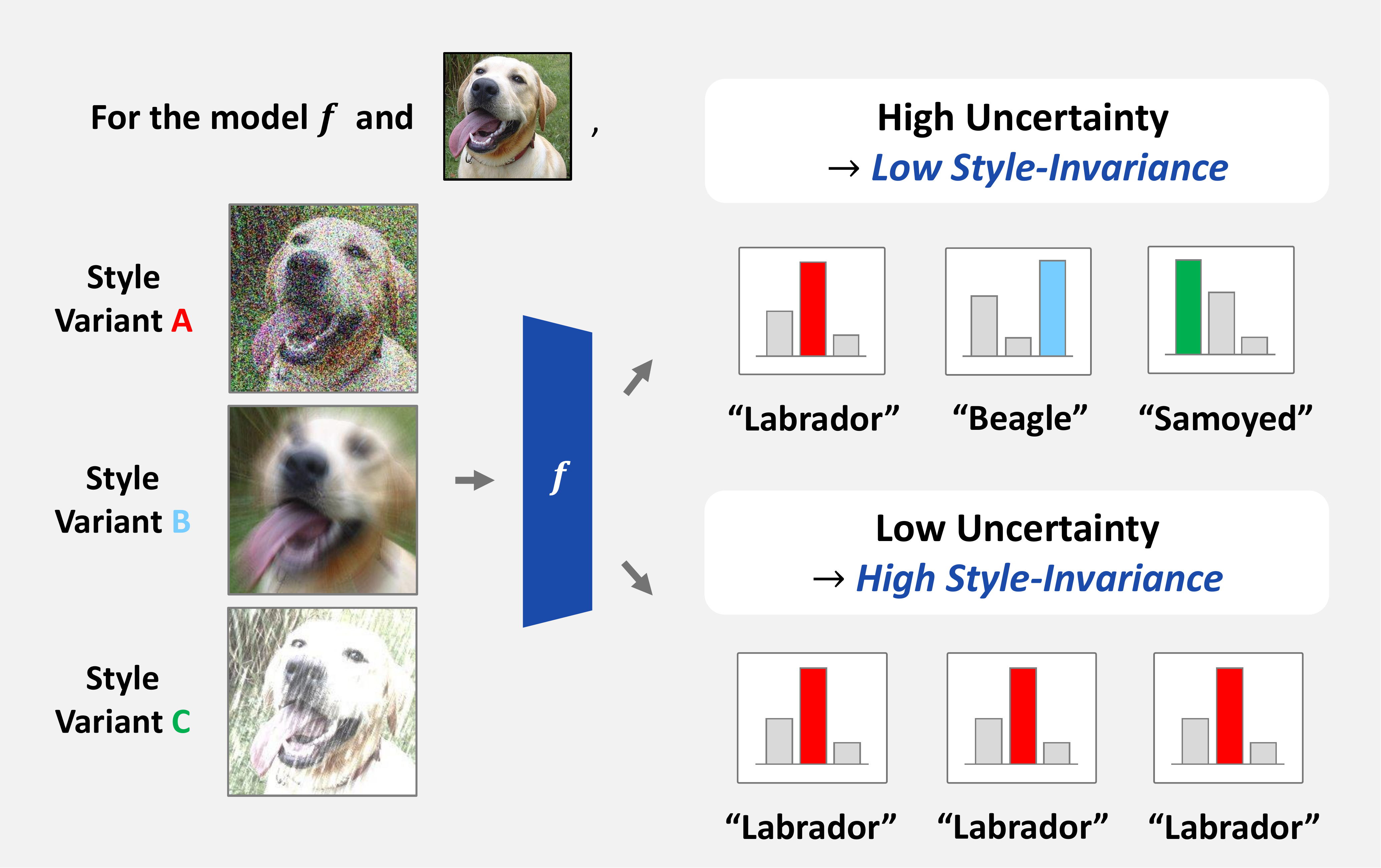} 
    \caption{Conceptual illustration of SICL's idea. SICL estimates correctness likelihood by assessing style invariance. Consistent predictions across style variants (generated by feature perturbation) (bottom) indicate high confidence in correctness, while low style invariance (top) indicate high uncertainty and lower confidence.}
    \label{Fig:overview}
    \vspace{-10pt}
\end{figure}

Test-time adaptation (TTA) has emerged as a promising solution to this challenge, assuming a realistic scenario of model deployment - where a deployed model aims to continuously adapt to unseen distribution using unlabeled test data. Thus, recent advances in TTA~\cite{cotta, rotta, sotta, sar, note, yuan2023robust} have increasingly sought to tackle realistic distribution patterns during model deployment, including temporal correlations~\cite{note} and out-of-distribution (OOD) samples~\cite{sotta}. However, a critical aspect in model deployment — ensuring the reliability of model predictions — remains largely unexplored. Consequently, this limits the applicability of TTA methods in risk-sensitive domains such as autonomous driving~\cite{segu2023darth, li2023test}, and medical application~\cite{he2021autoencoder, yang2022dltta, valanarasu2024fly}.

We observe that prior calibration methods, including \textit{temperature-scaling} approaches~\cite{guo2017calibration, wang2020transferable, hupseudo, park2020calibrated} and \textit{ensembling} approaches~\cite{gal2016dropout, lakshminarayanan2017simple}, assume a fixed model and static test distribution, consequently failing in dynamic regime of TTA. Specifically, we find that during TTA: (1) models exhibit varying calibration patterns when faced with varying test distributions (Section~\ref{sec:difficulty-calibration}), and (2) the self-supervised (\textit{e.g.} entropy-minimization) objective utilized in TTA methods~\cite{sotta, rotta, sar, tent} progressively induces overconfidence, undermining the reliability of predictions (Section~\ref{sec:importance-realistic}). This underscores the necessity for a dynamic calibration approach designed to adapt to these evolving conditions.

In light of these challenges, we propose \textit{Style Invariance as a Correctness Likelihood (SICL)}, a robust calibration framework designed for real-world TTA scenarios. Grounded in the causal assumption~\cite{mitrovic2020representation,von2021self} that predictions should depend solely on content rather than style (Section ~\ref{sec:background}), SICL estimates confidence by measuring the consistency between the original prediction and those obtained from style-shifted variants. By further employing relaxation weights based on the variance of style-induced predictions, SICL ensures a robust, instance-specific calibration across diverse, unseen distributions. Notably, SICL is backpropagation-free, making it easily integrable with various TTA baselines.

We evaluate SICL against state-of-the-art calibration baselines~\cite{guo2017calibration, gal2016dropout, wang2020transferable, hupseudo}, upon three widely used benchmarks, namely CIFAR10-C, CIFAR100-C, and ImageNet-C, across two practical scenarios with three model architecture. Our results suggest that SICL consistently outperforms prior calibration methods by a large margin, showing up to 13\%p reduction  in expected calibration errors (ECE) compared to existing calibration methods under all experimental setups (Section \ref{sec:main_results}). Our extensive qualitative analysis validates the \textit{content-preserving} and \textit{style-varying} characteristics of SICL (Section~\ref{sec:analysis}), while our ablative studies demonstrates the effectiveness of each of our components.

\vspace{+0.05in}
We summarize our contributions as follows.
\vspace{+0.05in}
\begin{itemize}

\item
We shed light on the previously underexplored challenges and importance of uncertainty estimation in TTA scenarios, emphasizing the unique complexities posed by evolving models and distributions at test-time.
\item  
We introduce SICL, the first framework to leverage style invariance for model calibration, demonstrating robust performance under diverse realistic shifts without usage of both source data and target labels.
\item 
Our method achieves state-of-the-art calibration performance with a large margin across real-world scenarios including dynamic distribution shifts while being seamlessly applicable to various TTA methods.

\end{itemize}

\definecolor{lightcyan}{RGB}{224, 240, 255} 
\definecolor{headercyan}{RGB}{200, 220, 240} 
\definecolor{verylightblue}{RGB}{235, 245, 255} 
\definecolor{verylightred}{RGB}{255, 235, 235} 
\newcommand{\cmark}{\checkmark} 
\newcommand{\xmark}{\textbf{\texttimes}} 
\section{Related Work}\label{sec:related_work}

\paragraph{Uncertainty Calibration.} 
The goal of calibration is to align the confidence of a model with its actual accuracy, ensuring that the probability estimates reflect the true predictive uncertainty. Existing methods~\cite{guo2017calibration, liu2022devil, ding2021local, balanya2024adaptive, yu2022robust} typically exploit temperature scaling to adjust the logit. \textit{Temperature-scaling methods} assume a fixed model and distribution, scaling the entire data with a single temperature. As a result, they fail to address the varying characteristics of each test instance under various distribution changes and changing models. Moreover, obtaining the temperature often requires additional training with labeled data~\cite{guo2017calibration, wang2020transferable, park2020calibrated}, or pseudo-data~\cite{hupseudo}, which are unavailable at test time and adds a burden on top of the adaptation process. Meanwhile, \textit{ensembling methods} ~\cite{gal2016dropout, lakshminarayanan2017simple} use dropout or model ensembling by averaging stochastic predictions (ensemble candidates) for more reliable confidence estimates. However, dropout often produces poor candidate representations due to its process of randomly disabling neurons of the model (Section \ref{sec:motivation}). Also, Deep Ensembles~\cite{lakshminarayanan2017simple} require training several models, which is undesirable in TTA. As illustrated in Table \ref{tab:calibration_comparison}, existing methods face several limitations when applied to TTA setup, underscoring the need for tailored calibration strategies for real-world TTA.

\begin{table}[t!]
    \centering
    \tiny 
    \renewcommand{\arraystretch}{1} 
    \setlength{\tabcolsep}{1pt} 
    \resizebox{\columnwidth}{!}{  
    \begin{tabular}{>{\centering\arraybackslash}p{1cm} |>{\centering\arraybackslash}p{1.2cm} >{\centering\arraybackslash}p{1cm} >{\centering\arraybackslash}p{1.2cm} >{\centering\arraybackslash}p{1.2cm} >{\centering\arraybackslash}p{1.2cm}}
        \toprule
        \textbf{Category} & \textbf{Calibration Method} & \textbf{No Source Data} & \textbf{No Extra Training} & \textbf{Instance-wise Calibration} & \textbf{Quality of Ensemble Candidate} \\
        \midrule
        \multirow{3}{*}{\makecell{Temperature \\Scaling}} & Vanilla TS  & \cellcolor{verylightred}$\xmark$ & \cellcolor{verylightred}$\xmark$& \cellcolor{verylightred}$\xmark$ & - \\
        & TransCal  & \cellcolor{verylightred}$\xmark$ & \cellcolor{verylightred}$\xmark$ & \cellcolor{verylightred}$\xmark$ & - \\
        & PseudoCal  & \cellcolor{verylightblue}$\cmark$& \cellcolor{verylightred}$\xmark$ & \cellcolor{verylightred}$\xmark$ & - \\
        \midrule
        \multirow{2}{*}{Ensembling} & MC Dropout  & \cellcolor{verylightblue}$\cmark$ & \cellcolor{verylightblue}$\cmark$ & \cellcolor{verylightblue}$\cmark$ & Low \\
        & \textbf{SICL (Ours)}  & \cellcolor{verylightblue}$\cmark$ & \cellcolor{verylightblue}$\cmark$ & \cellcolor{verylightblue}$\cmark$ & High \\
        \bottomrule
    \end{tabular}}
    \caption{Comparison of calibration methods.}
    \label{tab:calibration_comparison}
    \vspace{-10pt}
\end{table}

\paragraph{Test-Time Adaptation.} 
The goal of Test-Time Adaptation (TTA) is to adapt deployed models to distributional shifts encountered during inference without access to labeled target data. Typical TTA methods~\cite{tent,chen2022contrastive,liang2020we, memo} use techniques like entropy minimization, pseudo-labeling, and target-specific feature refinement to respond to these shifts. Recently, interest has surged in more practical TTA methods~\cite{sotta,rotta,cotta,note,sar,niu2023towards} designed for real-world test streams (e.g., continual shifts, noisy streams). However, a key limitation of existing TTA methods is their tendency toward overconfidence, leading to unreliable confidence estimates—a problem that becomes even more severe in dynamic test streams (Section \ref{sec:observation}). This unreliability poses significant risks in high-stakes decision making scenarios. These challenges highlight the need for robust uncertainty estimation in TTA to ensure reliability. To address these similar concerns, AETTA~\cite{Lee_2024_CVPR} introduced dropout to estimate accuracy. However, dropout generates low-quality ensemble candidates with distorted content, resulting in ineffective uncertainty estimation (Section \ref{sec:content_analysis}).

\begin{figure*}[t!]
    \centering
    \includegraphics[width=\linewidth]{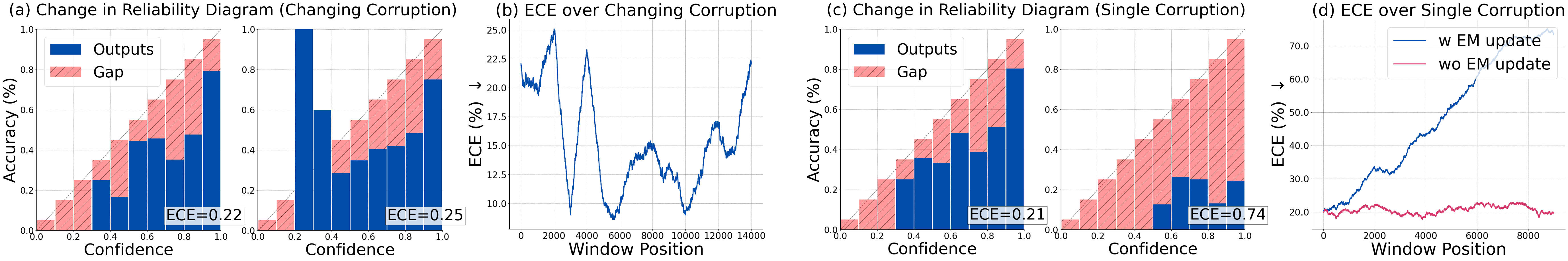}
    \caption{
    Observations upon TTA calibration patterns: (a) \& (b) show reliability and calibration error changes across temporal distribution shifts, while (c) \& (d) display these changes under a fixed Gaussian Noise corruption.
    }
    \label{Fig:observation}
    \vspace{-10pt}
\end{figure*}
\section{Preliminaries}\label{sec:background}


\paragraph{Style-Content Differentiation in Representation Learning}
We revisit the style-content assumptions of representation learning~\cite{mitrovic2020representation, von2021self} and ideal properties in an adapted model. The assumptions underpinning our research motivation (Section ~\ref{sec:motivation}) are as follows.
\begin{itemize}
    \item \textbf{Data Generation from Style and Content}: Data is generated from two distinct variables—content and style—where only content is pertinent to downstream tasks, while style introduces irrelevant variations.
    \item \textbf{Independence of Content and Style}: Content and style are independent variables, and style transformations do not alter the underlying content of the data.
    \item \textbf{Content-driven Prediction}: An ideal machine learning model makes its predictions based on content rather than style.
\end{itemize}
We believe this assumption is valid in most practical scenarios, as empirically evidenced in Section~\ref{sec:ablation}. Furthermore, in situations where the model cannot distinguish a sample’s content from its style completely (\textit{i.e.}, model collapse), addressing this issue requires adaptation rather than calibration, as calibration alone cannot correct the fundamental inaccuracies within the model.

\section{Challenges in Calibration under TTA} 

\label{sec:observation}
In this section, we examine the unique characteristics of calibration within the TTA framework. Unlike the typical environment addressed by conventional calibration approaches, TTA operates in a dynamic setting where the model continuously adapts to unseen distribution shifts without assuming a fixed test distribution, nor a fixed model state during calibration. Specifically, we find that in such dynamic settings, (1) we cannot expect consistent and reliable calibration performance in such dynamic distributions (Section~\ref{sec:difficulty-calibration}), and (2) most TTA methods~\cite{tent, sotta, sar, rotta} inadvertently push the model toward overconfidence as it adapts. (Section~\ref{sec:importance-realistic}).



\subsection{Calibration Instability in Test-Time Adaptation}\label{sec:difficulty-calibration}
Figure~\ref{Fig:observation} (a,b) demonstrates how distribution shifts affect calibration performance. We analyze calibration metrics while transitioning between 15 corruption types (\textit{i.e.}, Gaussian noise $\rightarrow$ Shot noise $\rightarrow$ Impulse noise ..) with a fixed model. The reliability diagrams for the initial and final 1,000 samples (Figure~\ref{Fig:observation} a) reveal markedly different patterns, despite similar Expected Calibration Error (ECE) values of approximately 20\%. Figure~\ref{Fig:observation} (b) tracks the expected calibration error using a sliding window analysis during corruption type transitions, where we find substantial fluctuations. These observations suggest that traditional calibration methods, which assume a static test distribution, will inadvertently yield unreliable performance when faced with dynamic distribution shifts, as their effectiveness becomes heavily dependent on the characteristics of the current input batch.

\subsection{Continuous Adaptation Causes Overconfidence}\label{sec:importance-realistic}

The impact of model adaptation on calibration performance is illustrated in Figure~\ref{Fig:observation} (c,d). During Test-Time Adaptation (TTA), models continuously adapt to novel distributions through online learning, primarily utilizing Entropy Minimization (EM) objectives~\cite{tent, sar, sotta, rotta}. However, EM-based adaptation can lead to excessive reliance on the model's internal predictions, resulting in overconfidence. Figure~\ref{Fig:observation} (c) compares reliability diagrams between the initial and final models after TENT~\cite{tent} adaptation on CIFAR-10C Gaussian noise, revealing a threefold increase in calibration errors due to heightened overconfidence. Figure~\ref{Fig:observation} (d) tracks this degradation through sliding window analysis, showing monotonic increases in calibration loss during EM-based updates. These findings demonstrate that entropy minimization, despite its prevalence in TTA methods, systematically drives models toward overconfidence, even under static test distributions. This is consistent with the finding of the Entropy Enigma~\cite{press2024entropy} that updates through EM ultimately fail.

These observations underscore the inherent challenges in achieving robust calibration in TTA, highlighting the need for tailored calibration approaches that can address evolving, unpredictable conditions without external supervision.



\section{Motivation}\label{sec:motivation}
\begin{figure}[t!]
    \centering
    \includegraphics[width=\columnwidth]{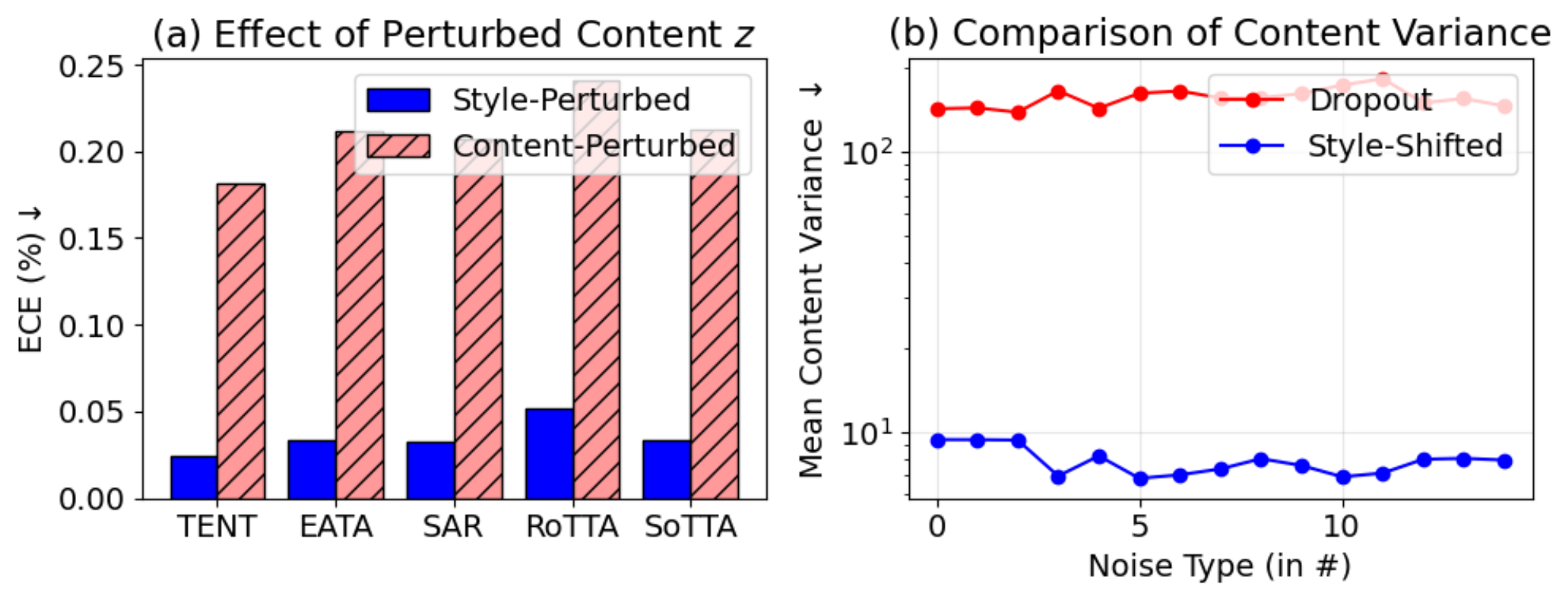}
    \vspace{-20pt}
    \caption{
    Motivative analysis for desirable ensemble candidates. From left to right, each visualize (a) calibration performance comparison between style and content perturbation, and (b) mean \textit{ContentVariance} of ensemble candidates over various corruption types.
    }
    \label{Fig:motivation}
    \vspace{-10pt}
\end{figure}

In the TTA scenario, the aforementioned challenges fundamentally arise because the distribution of the test stream can change very dynamically. In this respect, instance-wise calibration through \textit{ensembling} is more appropriate than \textit{temperature-scaling} approaches~\cite{guo2017calibration, wang2020transferable, park2020calibrated, hupseudo} that scale with a single temperature for a fixed distribution. Moreover, \textit{temperature-scaling} methods require additional data and training for obtaining the temperature parameter, which is undesirable in TTA. For \textit{ensemble-based} calibration, the quality of ensemble candidates is crucial for the uncertainty estimation. However, the quality of ensemble candidate representations remains suboptimal in state-of-the-art \textit{ensembling} approach, MC Dropout~\cite{dropout}.
To realize proper instance-wise calibration with ensembling, confidence should be estimated through consistency with ensemble candidates where only the style is changed while preserving the content defining the instance itself. However, dropout could generate poor candidate representations since the process of randomly disabling neurons, which often includes informative nodes, results in representations that lose critical content information.

Here, we provide a motivative analysis of the candidate embeddings $\mathbf{z}_i$ of the target image instance $\mathbf{x}_i$ from CIFAR10-C. First, we provide insights on the necessity of preserving the content for candidate representation. Upon this, we offer comparison for candidate representation  produced by MC Dropout~\cite{dropout} and style-shifted representation from MixStyle~\cite{zhou2021domain}, a representative style-variation method.

\begin{figure*}[t!]
    \centering
    \includegraphics[width=.8\linewidth]{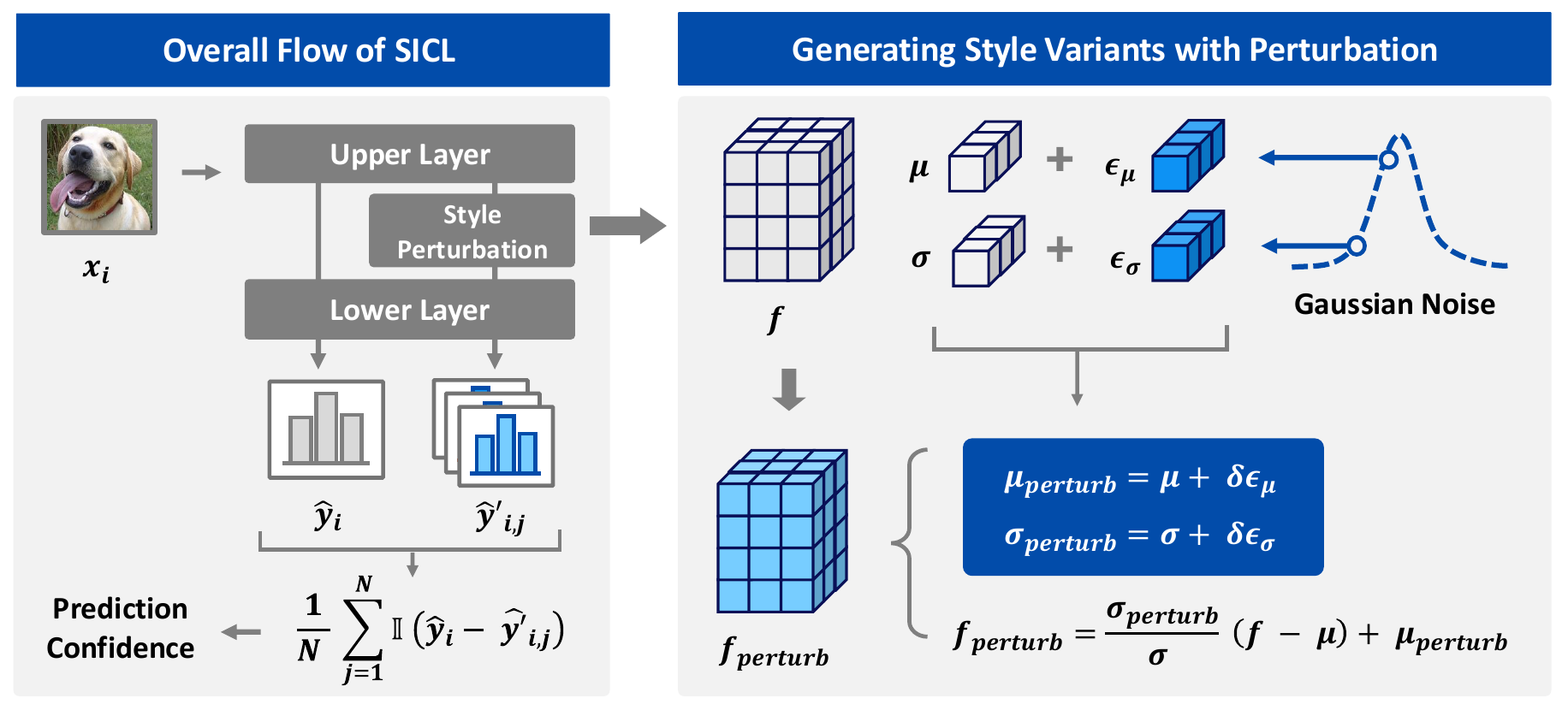}
    \caption{
    Detailed illustration of the SICL framework, demonstrating how style invariance is leveraged to estimate correctness likelihood. Given an original sample $\mathbf{x}_i$, SICL generates multiple style-shifted variants as ensemble candidates through feature-level style perturbations. Prediction consistency across these variants is used as a final prediction confidence, where high style-invariance indicates a higher confidence.
    }
    \label{Fig:main_figure}
    \vspace{-10pt}
\end{figure*}


\subsection{Ideal Characteristics of Ensemble Candidates}\label{sec:ideal_ensemble}

Here, we empirically validate our assumption that optimal ensemble candidates should be (1) style-variant and (2) content-preserving. Figure~\ref{Fig:motivation} (a) presents our results.
We visualize the effect of ensembling 20 different candidates generated via (1) \textit{Style Perturbation}, adding gaussian noise into style statistics (channel-wise mean $\mu_i$ and standard deviation $\sigma_i$ of earlier layer feature $\mathbf{z}_i$ of ResNet, following the seminal style-transfer works~\cite{huang2017arbitrary, dumoulin2017learned, Gatys_2016_CVPR, ulyanov2016instance}) and (2) \textit{Content Perturbation}, appling perturbations to the embeddings' content rather than style. Specifically, we add Gaussian noise to whitened features $\frac{\mathbf{f}_i - \mu_i}{\sigma_i}$ and then restoring them to $\mu_i$ and $\sigma_i$. Building on prior studies~\cite{huang2017arbitrary, pan2018two} that identify whitened features as a content representation, our results show that content corruption leads to severe calibration errors, over 4 times higher than with style perturbations, highlighting the importance of separating style to preserve the content during the ensemble candidate generation process.

\subsection{Embedding-level Comparative Analysis}\label{sec:content_analysis}

Based on the intuition that a candidate embedding that well-preserves its content shall retain their geometric relationship to the embedding of its class centroid, we compare style-shifted embeddings with those generated by MC Dropout via Mahalanobis distance. Specifically, for each sample $\mathbf{x}_i$ and its $N$ ensemble candidates $\mathbf{z'}_{i,j}$, we compute the Mahalanobis distance $\mathcal{D}_M(\mathbf{z}_i) = \sqrt{(\mathbf{z}_i - \boldsymbol{\mu}_c)^T \mathbf{\Sigma}_c^{-1} (\mathbf{z}_i - \boldsymbol{\mu}_c)}$ between its original embedding $\mathbf{z}_i$ and the corresponding class centroid $\boldsymbol{\mu}_c$. Here, we define \textit{ContentVariance} of candidate embedding $\mathbf{z'}_{i}$ as follows:

\[
\textit{ContentVariance}(\mathbf{z'}_i)= \left| \mathcal{D}_M\left(\mathbf{z}_i\right) - \frac{1}{N} \sum_{j=1}^{N}{\mathcal{D}}_M\left(\mathbf{z'}_{i,j}\right) \right|
\]
Note that, through Mahalanobis distance, we account for the covariance structure in the embedding space—leading to a more reliable measure of how well the candidate embeddings preserve content-related features. Figure~\ref{Fig:motivation}(b) shows the results of average \textit{ContentVariance} for all samples in CIFAR10-C. Across all corruptions, style-shifted embeddings achieve substantially lower \textit{ContentVariance} than MC Dropout embeddings, highlighting their superior ability to maintain the original content information.

\section{Methodology}\label{sec:method}

Through the analysis in Section \ref{sec:motivation}, we recognize that the current \textit{ensemble-based} calibration methods represented by MC Dropout are insufficient to solve the challenges presented in Section \ref{sec:observation}. Therefore, we propose a methodology that generates better ensemble candidate representations based on the style-content assumptions in Section \ref{sec:background}. Building on these assumptions, we present that style invariance can serve as a robust estimator of correctness likelihood by proposing SICL in scenarios lacking label information from target domains. The overall framework of SICL is visualized in Figure \ref{Fig:main_figure}. In the following subsection, we describe the SICL by outlining the overall procedure (Section \ref{sec:method-sicl}), the generation of style variants through feature perturbation (Section \ref{sec:method-style variation}), and a relaxation mechanism to address overconfidence during model collapse (Section \ref{sec:method-relaxation}).

\subsection{Style Invariance as a Correctness Likelihood}\label{sec:method-sicl}


Given an original sample \( \mathbf{x}_i \), its feature  \( \mathbf{z}_i \) and a trained model \( f \), we calculate the final prediction confidence by measuring instance-wise style invariance, with an additional relaxation term to handle non-ideal cases. To do this, we perform the following steps:



\begin{enumerate}
    \item \textbf{Generating Style Variants}: Create a set of \( N \) style-transformed samples \( \{ \mathbf{z}_{i,j}^{s\prime} \}_{j=1}^{N} \) as ensemble candidates
    by applying different style transformation \( T \) with style \( s_j \) to the \( \mathbf{z}_i \), resulting in 
   \( \mathbf{z}^{s\prime}_{i,j} = T(\mathbf{z}_i, s_j), \quad \forall j \in \{1, \dots, N\}. \)
    The style transformation \( T \) alters the style while preserving the content.

    \item \textbf{Measuring Style Invariance as a Initial Confidence}: Calculate the style invariance ratio \( \gamma^{style}_i \) as initial prediction confidence $p_{\text{init.}}(y_i)$ with the prediction consistency between the original prediction \( \hat{y_i} = f(\mathbf{x}_i) \) and the predictions on the style variants \( \{ y_{i,j}^{s\prime} \}_{j=1}^{N} \). Concretely,
    \[
    p_{\text{init.}}(y_i) = \gamma^{style}_i = \frac{1}{N} \sum_{j=1}^{N} \mathbb{I}\left( \hat{y}_i = \hat{y}^{s\prime}_{i,j} \right)
    \]
    where \( \mathbb{I} \) is the indicator function.

    \item \textbf{Handling Potential Collapse with Relaxation}: We derive $p_{\text{init.}}(y_i)$ based on the assumption that models perform content-driven prediction. However, this may not always hold in practical scenarios. To address this, we compute the final calibrated confidence by applying relaxation $\omega_\text{relaxation}$, which mitigates the overconfidence arising from the model collapses. This can be formulated as follows:
    \[
    p_{\text{calibrated}}(y_i) = \omega_\text{relaxation}\cdot p_{\text{init.}}(y_i)
    \]

\end{enumerate}

\subsection{Style Variation through Perturbation}\label{sec:method-style variation}

Key component for measuring instance-wise style invariance is securing a style transformation \( T \) that is able to simulate realistic shift by preserving the semantic content information of instance while altering its style.  To achieve this, we extract style representation from feature statistics and generate style-variants by perturbing this representation with a Gaussian noise. We exploit the channel-wise mean $\mu$ and standard deviation $\sigma$ of a feature map as a style representation, following the seminal style-transfer works~\cite{huang2017arbitrary, dumoulin2017learned, Gatys_2016_CVPR, ulyanov2016instance}. Based on these studies demonstrating that shallow layers tend to capture significant style information, we utilize the feature representation from the first layer.
 
 For each \(\mathbf{x}_i\), we compute the channel-wise mean and standard deviation of the feature map \(\mathbf{f}_i \in \mathbb{R}^{ C \times H \times W} \) as:
\begin{equation}
\begin{aligned}
   \mu_i = \frac{1}{H \times W} \sum_{h=1}^{H} \sum_{w=1}^{W} \mathbf{f}_i^{(c, h, w)},
\end{aligned}
\end{equation}

\begin{equation}
\begin{aligned}
\sigma_i = \sqrt{\frac{1}{H \times W} \sum_{h=1}^{H} \sum_{w=1}^{W} \left(\mathbf{f}_i^{(c, h, w)} - \mu_i\right)^2.}
\end{aligned}
\end{equation}
To realize style perturbation, we add random noise to the original \(\mu_i\) and \(\sigma_i\) or apply modified values. The perturbed mean \(\mu_{\text{perturb}}\) and standard deviation \(\sigma_{\text{perturb}}\) are defined as:
\begin{equation}
\begin{aligned}
   \mu_{\text{perturb}} = \mu_i + \delta\epsilon_\mu, \quad \sigma_{\text{perturb}} = \sigma_i + \delta\epsilon_\sigma, 
\end{aligned}
\end{equation}
where \(\epsilon_\mu\) and \(\epsilon_\sigma  \sim \mathcal{N}(0, 1)\) are random noise which follow the standard Gaussian distribution. Here, \(\delta\) is a standard deviation of $\mu_i$ for online test batch. While a similar approach like MixStyle~\cite{zhou2021domain}, which synthesizes a new style by mixing the style statistics of two samples, it is limited by its dependency on the input batch - which is undesirable in a TTA scenario. In contrast, leveraging stochastic perturbation with Gaussian noise enables the generation of diverse style representations with varying directions and intensities, offering greater flexibility and variation (Section \ref{sec:analysis}).

Using the \(\mu_{\text{perturb}}\) and \(\sigma_{\text{perturb}}\), we generate perturbed features as style variants, denoted by  \(\mathbf{f}_{\text{perturb}}\):
\begin{equation}
\begin{aligned}
   \mathbf{f}_{\text{perturb}} = \sigma_{\text{perturb}} \cdot \frac{\mathbf{f}_i - \mu_i}{\sigma_i} + \mu_{\text{perturb}}.
\end{aligned}
\end{equation}
We use the consistency between the predictions of the 
\( N \) style variants generated from \( \mathbf{f}_{\text{perturb}} \) and the original sample's prediction as the confidence of original sample \( \mathbf{x}_i \).


\subsection{Relaxation for Collapse}\label{sec:method-relaxation}
SICL assumes that a model effectively disentangles style and content. However, during continuous adaptation in TTA, model often falls into a collapsed state where it overly attends to the style rather than content during its predictions, hence making the style invariance term unreliable. Thus, we introduce a relaxation weight $\omega_\text{relaxation}$ that penalizes content-independent predictions that overly attend on the \textit{style}.
To identify the content-independent predictions, we calculate the content invariance ratio $\gamma^{content}_i$ which quantifies the invariance of predictions with respect to content distortion. Specifically, we calculate the content invariance ratio $\gamma^{content}_i$ as follows:
\[
\gamma^{content}_i = \frac{1}{N} \sum_{j=1}^{N} \mathbb{I}\left( \hat{y}_i = \hat{y}^{c\prime}_{i,j} \right)
\]
where $\hat{y}^{c\prime}_{i,j}$ denotes the predictions of content-distorted sample $\mathbf{z}^{c\prime}_i$. Grounded in prior works~\cite{huang2017arbitrary, pan2018two}, we obtain the whitened embeddings (content-specific embedding) to simulate content distortion  as follows: $\mathbf{f}_{\text{white}} = \frac{\mathbf{f}_i - \mu_i}{\sigma_i + \epsilon}$. Next, we add gaussian noise to perturb its content; concretely as follows:
\[
    \mathbf{f}_{\text{content-perturb}} = \sigma_i \cdot (\mathbf{f}_{\text{white}} + \mathbf{\epsilon}) + \mu_i 
\]
where, $ \mathbf{\epsilon}$ is the gaussian noise \(\epsilon_\sigma  \sim \mathcal{N}(0, \sigma^{\text{white}}_i \cdot I)\). In essence, we're applying standard gaussian noise scaled to the channel-wise standard deviations of the whitened features.
The final relaxation weight is defined as $\omega_\text{relaxation} = 1-\gamma^{content}_i$, which effectively penalizes predictions that remain unchanged despite content modifications.

\section{Experiments}\label{sec:experiments}
In this section, we present the experimental setup along with both quantitative and qualitative results. Additional details on the experimental configurations and outcomes are provided in the Appendix. 
\subsection{Experimental Setup}\label{sec:exp_setup}
\paragraph{Datasets \& Models.}
We evaluate SICL on three widely-used benchmarks: CIFAR10-C, CIFAR100-C, ImageNet-C~\cite{cifarc}. These datasets simulate diverse types of distribution shifts, from simple noise to complete domain shifts, commonly encountered in real-world scenarios. CIFAR10-C and CIFAR100-C contains 15 different synthetic corruptions with five levels of corruption, where we adopt the highest severity level 5 for our experiments. CIFAR10-C/CIFAR100-C contain 10/100 classes, with 10000/1000 samples, respectively. ImageNet-C
dataset contain 50,000 test samples with 1000 classes. For backbone architecture, We use pretrained ResNet-50, 101~\cite{resnet} and ViT~\cite{dosovitskiy2020image} (ViT-Small for CIFAR10/100-C, and ViT-Base for ImageNet-C). For Vision Transformers, we implement our style perturbation approach on the patch embedding space by utilizing the patch-wise mean and standard deviation as a style statistics, following intuitions from TFS-ViT~\cite{noori2024tfs}.

\paragraph{Scenario.}
To account for diverse distributions in real-world test streams, we define the following two test stream distributions. 
\begin{itemize}
    \item \textbf{Benign}: Assuming a fixed test distribution where a single corruption type (\textit{e.g.} Gaussian noise) is fixed throughout test, where samples arrive in an \textit{i.i.d.} manner.
    \item \textbf{Dynamic}: Assuming a realistic test distribution, where corruption type has temporal correlation (\textit{i.e.} corruption types arriving at varying lengths). We adopt Dirichlet function to simulate temporal correlation of corruptions  (\textit{e.g. Gaussian noise $\rightarrow$ Shot noise $\rightarrow$ Defocus blur ...}).
\end{itemize}

\paragraph{Baselines.}
We evaluate SICL upon 5 different TTA methods - namely TENT~\cite{tent}, EATA~\cite{eata}, SAR~\cite{sar}, RoTTA~\cite{rotta} and SoTTA~\cite{sotta}. For ViT backbones, we consider TENT-LN and SAR-LN, both a variants which updates the affine parameters of the layer normalization layers instead of batch normalization. For calibration baselines, we consider 4 representative methods applicable during TTA scenario. Vanilla TS~\cite{guo2017calibration} applies a fixed temperature learned from the validation dataset of source to calibrate confidence. MC Dropout~\cite{gal2016dropout} uses stochastic dropout at inference to create an ensemble of predictions, capturing uncertainty through averaging. TransCal~\cite{wang2020transferable} estimates target calibration error with the source distribution by introducing importance weighting. Lastly, PseudoCal~\cite{hupseudo} trains the temperature parameter with pseudo-data generated by Mixup~\cite{zhang2018mixup}. Note that, other than MC Dropout, PseudoCal and SICL, we relax the constraints of TTA; allowing the usage of source data for calibrations.


\definecolor{verylightblue}{RGB}{235, 245, 255} 
\definecolor{verylightred}{RGB}{255, 235, 235} 

\begin{table*}[t]
\centering
\caption{Expected Calibration Error (ECE) (\%) of uncertainty estimation on CIFAR10-C, CIFAR100-C, and ImageNet-C across five TTA methods under benign and dynamic test streams. \textbf{Bold} numbers represent the lowest error.}
\label{tab:main_results}
\resizebox{\textwidth}{!}{%
\begin{tabular}{lllccccccccccccc} 
\toprule
\multicolumn{1}{c}{\textbf{Stream}} & \multicolumn{1}{c}{\textbf{Dataset}} & \multicolumn{1}{c}{\textbf{Method}} 
& \multicolumn{5}{c}{\textbf{ResNet-50}} 
& \multicolumn{5}{c}{\textbf{ResNet-101}} 
& \multicolumn{2}{c}{\textbf{ViT}} 
& \multicolumn{1}{c}{\textbf{Avg.}}\\
\cmidrule(lr){4-8}\cmidrule(lr){9-13}\cmidrule(lr){14-15}  
& & & \textbf{TENT} 
& \textbf{EATA} 
& \textbf{SAR} 
& \textbf{RoTTA} 
& \textbf{SoTTA} 
& \textbf{TENT} 
& \textbf{EATA} 
& \textbf{SAR} 
& \textbf{RoTTA} 
& \textbf{SoTTA} 
& \textbf{TENT} 
& \textbf{SAR} 
& \multicolumn{1}{c}{} 
\\
\midrule
\multirow{18}{*}{\textbf{Benign}} 
  & \multirow{6}{*}{\textbf{CIFAR10-C}} 
    & Vanilla TS  & 22.7 & 22.3 & 22.2 & 22.9 & 22.8 & 32.0 & 21.2 & 21.8 & 21.7 & 21.8 & 17.2 & 15.2 & {21.9} \\
  & 
    & MC Dropout  & 16.3 & 11.1 & 10.2 & 10.0 & 10.4 & 25.9 & 10.2 & 10.1 & 9.2  & 10.2 & 22.1 & 19.3 & 13.8 \\
  & 
    & TransCal    & 56.7 & 56.2 & 56.9 & 11.0 & 56.1 & 48.0 & 57.0 & 57.0 & 52.0 & 56.9 & 26.7 & 26.5 & 44.8 \\
  & 
    & PseudoCal   & 9.4  & 7.8  & \textbf{5.2}  & 7.9  & 8.4  & 18.4  & 7.9  & \textbf{5.4}  & 9.0  & 8.6  & 34.9  & 33.6 & 13.9 \\
  & 
    & \cellcolor{verylightblue}\textbf{SICL (Ours)} 
                  & \cellcolor{verylightblue}\textbf{6.4}  
                  & \cellcolor{verylightblue}\textbf{6.9}  
                  & \cellcolor{verylightblue}7.5  
                  & \cellcolor{verylightblue}\textbf{7.1}  
                  & \cellcolor{verylightblue}\textbf{7.5}  
                  & \cellcolor{verylightblue}\textbf{6.6}  
                  & \cellcolor{verylightblue}\textbf{7.6}  
                  & \cellcolor{verylightblue}8.0  
                  & \cellcolor{verylightblue}\textbf{7.3}  
                  & \cellcolor{verylightblue}\textbf{8.0}  
                  & \cellcolor{verylightblue}\textbf{8.4}  
                  & \cellcolor{verylightblue}\textbf{8.7}  
                  & \cellcolor{verylightblue}\textbf{7.7} \\
\cmidrule(lr){3-16}
  & 
    & Acc         & 77.1 & 75.7 & 77.5 & 74.9 & 77.0 & 68.4 & 76.8 & 78.2 & 76.3 & 78.2 & 53.0 & 53.0 &  \\
\cmidrule(lr){2-16}
  & \multirow{6}{*}{\textbf{CIFAR100-C}} 
    & Vanilla TS  & 9.6  & 22.7 & 7.0  & 12.5 & 6.0  & 13.8 & 36.4 & 9.3  & 9.6  & 7.9  & 18.2 & 17.2 & 13.8 \\
  & 
    & MC Dropout  & 13.5 & 16.0 & 8.7  & 13.6 & 8.4  & 20.0 & 12.7 & 9.2  & 8.6  & 8.5  & 19.9 & 18.2 & 12.8 \\
  & 
    & TransCal    & 50.0 & 27.5 & 8.1  & 36.6 & 12.0 & 18.7 & 13.2 & 56.3 & 11.7 & 23.8 & 23.0 & 22.5 & 22.6 \\
  & 
    & PseudoCal   & \textbf{7.4} & 7.3  & 8.7  & 8.1  & 9.0  & 10.8 & 7.4  & \textbf{8.4} & 7.5  & 6.6  & 34.9 & 33.6 & 13.5 \\
  & 
    & \cellcolor{verylightblue}\textbf{SICL (Ours)} 
                  & \cellcolor{verylightblue}7.5  
                  & \cellcolor{verylightblue}\textbf{6.6}  
                  & \cellcolor{verylightblue}\textbf{7.9}  
                  & \cellcolor{verylightblue}\textbf{6.2}  
                  & \cellcolor{verylightblue}\textbf{7.7}  
                  & \cellcolor{verylightblue}\textbf{5.4}  
                  & \cellcolor{verylightblue}\textbf{5.6}  
                  & \cellcolor{verylightblue}6.6  
                  & \cellcolor{verylightblue}\textbf{5.3}  
                  & \cellcolor{verylightblue}\textbf{6.6}  
                  & \cellcolor{verylightblue}\textbf{8.4}  
                  & \cellcolor{verylightblue}\textbf{8.7}  
                  & \cellcolor{verylightblue}\textbf{6.8} \\
\cmidrule(lr){3-16}
  & 
    & Acc         & 52.7 & 33.1 & 54.2 & 49.7 & 51.1 & 44.1 & 19.4 & 55.8 & 51.6 & 52.5 & 50.8 & 50.9 &  \\
\cmidrule(lr){2-16}
  & \multirow{6}{*}{\textbf{ImageNet-C}} 
    & Vanilla TS  & 23.9 & 23.2 & 13.4 & 13.4 & 13.2 & 20.7 & 36.4 & 15.7 & 13.7 & 15.7 & 19.8 & 11.9 & 17.4 \\
  & 
    & MC Dropout  & 11.0 & 13.7 & 15.2 & 14.6 & 16.8 & 12.0 & 12.7 & 13.6 & 11.8 & 12.6 & 28.9 & 28.8 & 16.7 \\
  & 
    & TransCal    & 14.5 & 8.6  & \textbf{8.1}  & 9.2  & 9.8  & \textbf{11.0} & 34.3 & 19.2 & 10.6 & 11.1 & 40.5 & 41.8 & 19.6 \\
  & 
    & PseudoCal   & 14.1 & 8.5  & 14.9 & 13.7 & 14.9 & 14.0 & 11.6 & 14.7 & 12.9 & 14.8 & 19.0 & 17.4 & 14.8 \\
  & 
    & \cellcolor{verylightblue}\textbf{SICL (Ours)} 
                  & \cellcolor{verylightblue}\textbf{10.0}  
                  & \cellcolor{verylightblue}\textbf{7.4}  
                  & \cellcolor{verylightblue}9.1  
                  & \cellcolor{verylightblue}\textbf{7.2}  
                  & \cellcolor{verylightblue}\textbf{8.4}  
                  & \cellcolor{verylightblue}13.6  
                  & \cellcolor{verylightblue}\textbf{9.6}  
                  & \cellcolor{verylightblue}\textbf{12.4}  
                  & \cellcolor{verylightblue}\textbf{9.2}  
                  & \cellcolor{verylightblue}\textbf{10.9}  
                  & \cellcolor{verylightblue}\textbf{15.5}  
                  & \cellcolor{verylightblue}\textbf{11.1}  
                  & \cellcolor{verylightblue}\textbf{10.7} \\
\cmidrule(lr){3-16}
  & 
    & Acc         & 36.3 & 21.0 & 33.2 & 31.1 & 29.0 & 40.0 & 22.9 & 36.3 & 32.7 & 30.5 & 40.7 & 49.0 &  \\
\midrule
\multirow{18}{*}{\textbf{Dynamic}} & \multirow{7}{*}{\textbf{CIFAR10-C}}
& Vanilla TS  & 12.9 & 18.4 & 23.7 & 21.3 & 23.6 & 11.4 & 18.5 & 23.7 & 21.6 & 23.7 & 11.2 & 10.1 & 18.3 \\
&  & MC Dropout  & 15.4 & 18.7 & 10.4 & 15.9 & 10.4 & 17.8 & 19.5 & 10.4 & 14.3 & 10.4 & 23.8 & 24.0 & 15.9 \\
&  & TransCal    & 56.4 & 47.3 & 57.0 & 43.8 & 56.9 & 53.1 & 46.2 & 57.2 & 46.2 & 57.2 & 28.1 & 27.7 & 48.1 \\
&  & PseudoCal   & 17.1 & 7.5  & 12.4 & 14.1 & 12.5 & 23.4 & 7.3 & 12.5 & 14.4 & 12.5 & 25.6 & 25.5 & 15.4 \\
&  & \cellcolor{verylightblue}\textbf{SICL (Ours)} & \cellcolor{verylightblue}\textbf{5.9}  & \cellcolor{verylightblue}\textbf{5.8}  & \cellcolor{verylightblue}\textbf{6.9}  &\cellcolor{verylightblue} \textbf{6.2}  & \cellcolor{verylightblue}\textbf{6.9}  & \cellcolor{verylightblue}\textbf{6.6}  & \cellcolor{verylightblue}\textbf{6.3}  & \cellcolor{verylightblue}\textbf{7.3}  & \cellcolor{verylightblue}\textbf{7.1}  & \cellcolor{verylightblue}\textbf{7.4}  & \cellcolor{verylightblue}\textbf{7.8}  & \cellcolor{verylightblue}\textbf{7.4}  & \cellcolor{verylightblue}\textbf{7.1}  \\
\cmidrule(lr){3-16}
&  & Acc & 77.2 & 68.6 & 77.6 & 67.9 & 77.6 & 74.3 & 70.2 & 78.3 & 69.8 & 78.3 & 56.8 & 56.4 &  \\\cmidrule(lr){2-16}

& \multirow{7}{*}{\textbf{CIFAR100-C}}
& Vanilla TS  & 51.7 & 55.3 & 46.6 & 53.8 & 48.3 & 60.5 & 54.4 & 45.2 & 53.0 & 47.1 & 9.3  & 7.8  & 44.4 \\
& & MC Dropout  & 11.1 & 19.4 & 9.5  & 16.3 & 10.0 & 18.3 & 22.3 & 6.6  & 10.4 & 6.8  & 8.7  & 9.1  & 12.4 \\
&  & TransCal    & 43.6 & 34.5 & 48.4 & 29.8 & 46.4 & 33.4 & 35.9 & 50.0 & 32.1 & 47.8 & 23.3 & 22.7 & 37.3 \\
&  & PseudoCal   & 21.2 & 9.7  & 16.3 & 19.4 & 16.3 & 30.6 & 8.9 & 15.8 & 18.2 & 15.0 & 31.0 & 31.6 & 19.5 \\
&  & \cellcolor{verylightblue}\textbf{SICL (Ours)} & \cellcolor{verylightblue}\textbf{6.6}  & \cellcolor{verylightblue}\textbf{4.4}  & \cellcolor{verylightblue}\textbf{7.8}  & \cellcolor{verylightblue}\textbf{3.8}  & \cellcolor{verylightblue}\textbf{8.0}  & \cellcolor{verylightblue}\textbf{4.6}  & \cellcolor{verylightblue}\textbf{4.5}  & \cellcolor{verylightblue}\textbf{6.6}  & \cellcolor{verylightblue}\textbf{3.5}  & \cellcolor{verylightblue}\textbf{6.5}  & \cellcolor{verylightblue}\textbf{4.5}  & \cellcolor{verylightblue}\textbf{4.7}  & \cellcolor{verylightblue}\textbf{5.5}  \\
\cmidrule(lr){3-16}
&  & Acc & 45.7 & 43.7 & 50.6 & 36.7 & 48.6 & 36.9 & 44.7 & 52.2 & 38.8 & 50.0 & 28.3 & 27.6 &  \\\cmidrule(lr){2-16}

& \multirow{7}{*}{\textbf{ImageNet-C}}
& Vanilla TS  & 23.9 & 14.2 & 13.4 & 13.4 & 13.2 & 20.7 & 9.5 & 15.7 & 13.7 & 15.7 & 19.8 & 11.9 & 15.4 \\
&  & MC Dropout  & 12.1 & 12.4 & 6.8  & 9.2 & 6.7  & 17.9 & 15.9  & 14.1  & 18.4  & 7.9  & 28.9 & 28.8 & 14.9 \\
&  & TransCal    & 14.5 & 29.5 & \textbf{8.1}  & 9.2  & 8.0  & \textbf{11.0} & 32.2 & 9.2  & 10.6 & 9.1  & 40.5 & 41.8 & 18.6 \\
&  & PseudoCal   & 10.7 & 8.1 & 15.0  & 13.5  & 14.9  & 11.5 & 8.7  & \textbf{7.2}  & 9.0 & 7.1  & 13.5  & 11.0  & 10.9 \\
&  & \cellcolor{verylightblue}\textbf{SICL (Ours)} & \cellcolor{verylightblue}\textbf{8.9}  & \cellcolor{verylightblue}\textbf{7.7}  & \cellcolor{verylightblue}8.7  & \cellcolor{verylightblue}\textbf{6.5}  & \cellcolor{verylightblue}\textbf{6.3}  & \cellcolor{verylightblue}12.4  & \cellcolor{verylightblue}\textbf{8.5}  & \cellcolor{verylightblue}8.3  & \cellcolor{verylightblue}\textbf{8.4}  & \cellcolor{verylightblue}\textbf{6.9}  & \cellcolor{verylightblue}\textbf{10.1}  & \cellcolor{verylightblue}\textbf{8.0}  & \cellcolor{verylightblue}\textbf{8.4}  \\
\cmidrule(lr){3-16}
&  & Acc & 37.5 & 31.9 & 36.7 & 34.1 & 31.2 & 33.8 & 34.0 & 32.4 & 31.6 & 33.2 & 44.1 & 45.2 &  \\
\bottomrule
\end{tabular}%
}
\end{table*}

\subsection{Main Results}\label{sec:main_results}

In this section, we present our main experimental results, highlighting the effectiveness of SICL across various TTA scenarios. All experiments in this subsection were conducted three times, with results averaged to ensure fair comparisons. Also, we reported the average results across 15 corruption types for all experiments.  Table \ref{tab:main_results} shows the results on three datasets with three model architecture under benign, and dynamic stream. SICL achieved state-of-the-art calibration performance in almost all settings (63/72). On average across all datasets, scenarios, model architectures, and TTA methods, SICL outperforms baselines by 13.0\%p. In particular, it outperforms baselines by an average of 15.6\%p in the Dynamic Stream and by an average of 10.4\%p in the benign Stream, which demonstrates the superiority of SICL in situations where the test distribution is not fixed but changes dynamically.



\subsection{Qualitative Analysis}\label{sec:analysis}
\begin{figure}[htbp]
    \centering
    \includegraphics[width=\columnwidth]{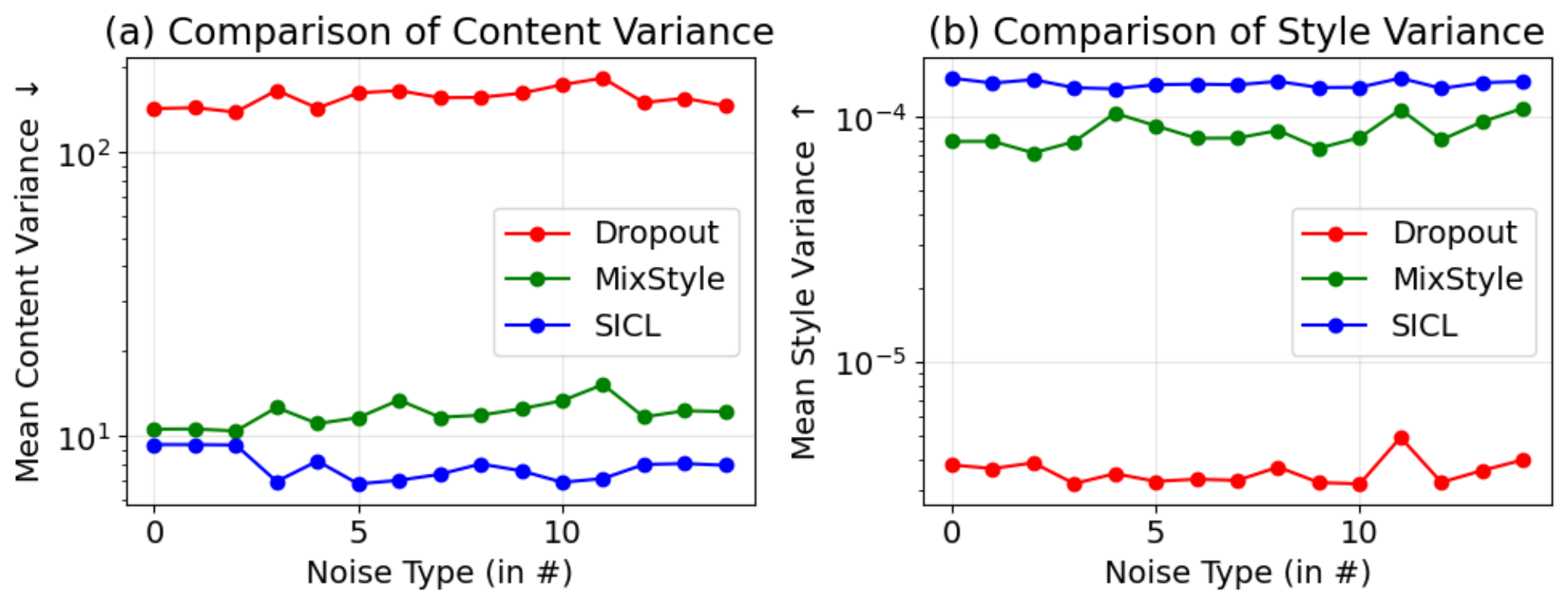}
    \caption{
    Qualitative analysis for style variants. From left to right, each visualize (a) mean \textit{ContentVariance} over various corruption types, and (b) mean \textit{StyleVariance} over various corruption types. 
    }
    \label{Fig:analysis}
    \vspace{-10pt}
\end{figure}
\begin{figure*}[t!]
    \centering
    \includegraphics[width=\linewidth]{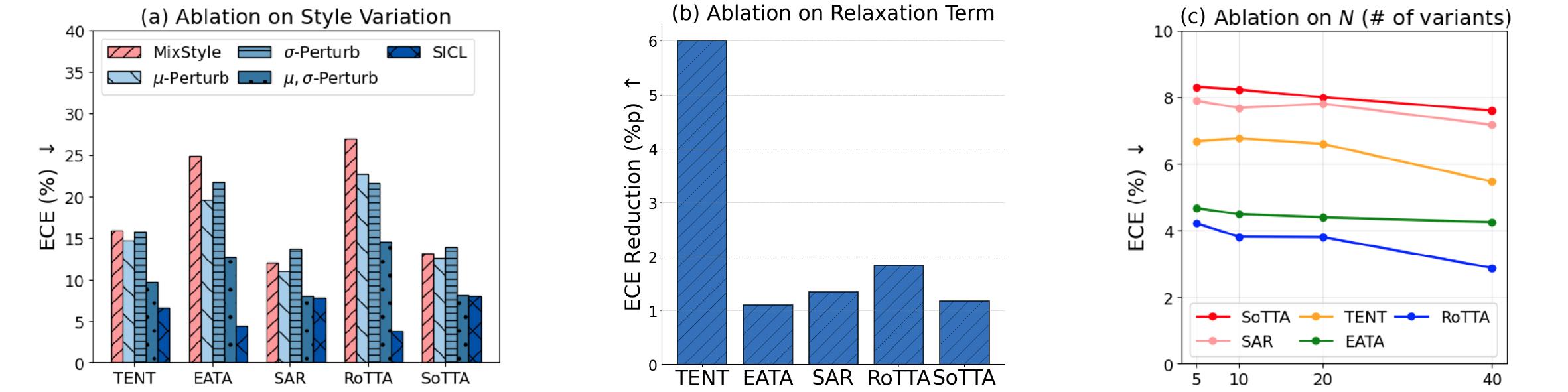}
    \caption{Ablation results on (a) individual components of SICL, (b) relaxation term - gain in expected calibration error via relaxation term, and (c) performances upon the number of variants $N$ across baseline TTA algorithms.}
    \label{fig:ablation_studies}
    \vspace{-10pt}
\end{figure*}In our earlier discussion (Section \ref{sec:motivation}), we examined the conditions for an ideal ensemble candidate representation, highlighting the importance of preserving the content of the original sample while simulating various style changes. Based on these conditions, we conducted an analysis to assess how qualitatively superior representations generated by our style perturbation are compared to conventional style-shift methods represented by MixStyle~\cite{zhou2021domain}. First, we measure the degree of content preservation by utilizing the average \textit{ContentVariance} defined in Section \ref{sec:content_analysis}. In Figure ~\ref{Fig:analysis} (a), we could see that our \textit{Style Perturbation} shows lower mean distance difference than MixStyle regardless of corruption types. Second, we measured style variance by repurposing the style reconstruction loss~\cite{johnson2016perceptual} to evaluate how well our method simulates diverse styles. For each sample $\mathbf{z}_i$ and its style-variant $\mathbf{z}^{s\prime}_i$, we calculate \textit{StyleVariance} as follows:
\[
 \textit{StyleVariance}(\mathbf{z}_i, \mathbf{z}^{s\prime}_i) =  \|G(\mathbf{z}_i) - G(\mathbf{z}^{s\prime}_i)\|_F^2 
\]
where $G$ denotes the Gram Matrix~\cite{Gatys_2016_CVPR} that encodes style information. In Figure~\ref{Fig:analysis} (b), we plot the average \textit{StyleVariance} of each sample in the CIFAR10-C dataset with its 20 style-variants. It can be seen that the style-perturbed embeddings from SICL exhibit a higher \textit{StyleVariance} across all corruption types than Dropout or MixStyle-generated embeddings. Thus, we find our style perturbation to generate proper candidate embeddings that 1) maintain content integrity while 2) exhibiting higher style diversity, enabling reliable and accurate prediction of model's confidence. In the case of MixStyle, it creates a new style by mixing the style statistics of two samples from the available target set (in the TTA situation, this is the online test batch). This is dependent on the composition of the input batch and cannot simulate as diverse a variation in the style space as \textit{Style Perturbation}, which leads to deficiencies in the quality aspect of style variant representation.


\subsection{Ablation Studies}\label{sec:ablation}
On the subsequent paragraphs, we conduct ablative studies on (1) effect of style variations by progressively removing each components, (2) effect of relaxation weight, and (3) effect of the number of variations generated ($N$). Note that all the ablative experiments were conducted upon CIFAR-100-C, with ResNet-50 backbone.

\paragraph{Ablation on style variation.}
We conduct a comprehensive ablation study to evaluate SICL against existing style variation methods and analyze the contribution of individual components. We mainly compare our \textit{Style Perturbation} with MixStyle~\cite{zhou2021domain}, which generates novel styles through pairwise mixing of feature-level style statistics within a batch. Additionally, we examine the individual impact of style components through: isolated perturbation of the fine-grained style attribute $\sigma$, and isolated perturbation of the coarse-grained style attribute $\mu$. The comparative results are presented in Figure~\ref{fig:ablation_studies} (a). It can be seen that our \textit{Style Perturbation} method consistently achieves superior performance across all TTA frameworks, validating its robustness and versatility. Notably, our analysis reveals that the influence of the coarse style attribute $\mu$ is greater than that of the fine style attribute $\sigma$, which aligns with prior findings~\cite{huang2017arbitrary}. Furthermore, the optimal calibration performance is achieved when combining both $\mu$ and $\sigma$, indicating their complementary roles in effective style variation.

\paragraph{Ablation on relaxation weight.}\label{sec:ablation_relaxation}
We further conduct an ablative study upon the relaxation term. The results are visualized in Figure~\ref{fig:ablation_studies} (b), where SICL denotes the final performance after applying relaxation weight  $\omega_\text{relaxation}$ to style invariance ratio $\gamma^{style}_i$ as a calibrated prediction confidence. It can be seen that $\omega_\text{relaxation}$ brings consistent improvements in uncertainty estimation performance, regardless of the TTA method. Here, we can see that the ECE reduction from relaxation is relatively small in SAR~\cite{sar} and SoTTA~\cite{sotta} compared to other methods, which can be attributed to the fact that these two methods already handle collapse to some extent through robust parameter updates based on Sharpness-Aware Minimization~\cite{foret2020sharpness}.



\paragraph{Hyperparameter sensitivity.} 
We show the hyperparameter sensitivity of SICL in Figure~\ref{fig:ablation_studies} (c). The reported expected calibration error is an average upon all data streams (\textit{i.e.} corruption type) and TTA methods. SICL doesn't have any hyperparameters other than the number of style variants. In Figure~\ref{fig:ablation_studies} (c), we plot the calibration performance with respect to number of style variants $N$. We found that SICL demonstrates robustness to the value of $N$ across all TTA methods. Specifically, it achieves sufficiently low calibration error (less than 10\% of ECE) with as few as $N=5$ style variants, and shows a consistent decrease in errors as $N$ increases. For all our experiments including main experiments and additional studies, we choose $N=20$.

\section{Conclusion}\label{sec:conclusion}
We tackle the critical yet underexplored problem of securing reliable uncertainty estimates in TTA scenarios. To address this challenge, we introduce SICL, the first framework to leverage style invariance for robust uncertainty estimation by providing instance-wise correctness likelihood in real-world test streams. Our comprehensive evaluation demonstrates that SICL consistently outperforms state-of-the-art calibration methods across diverse scenarios and settings while maintaining high practicality. Moreover, we believe our work could inspire further research in both uncertainty estimation for TTA and style-content-driven calibration.

{
    \small
    \bibliographystyle{ieeenat_fullname}
    \bibliography{main}
}
\maketitlesupplementary \appendix

\section{Additional Studies}

\begin{figure}[ht!]
    \centering
    \begin{subfigure}[t]{0.32\textwidth} 
        \centering
        \includegraphics[width=\textwidth]{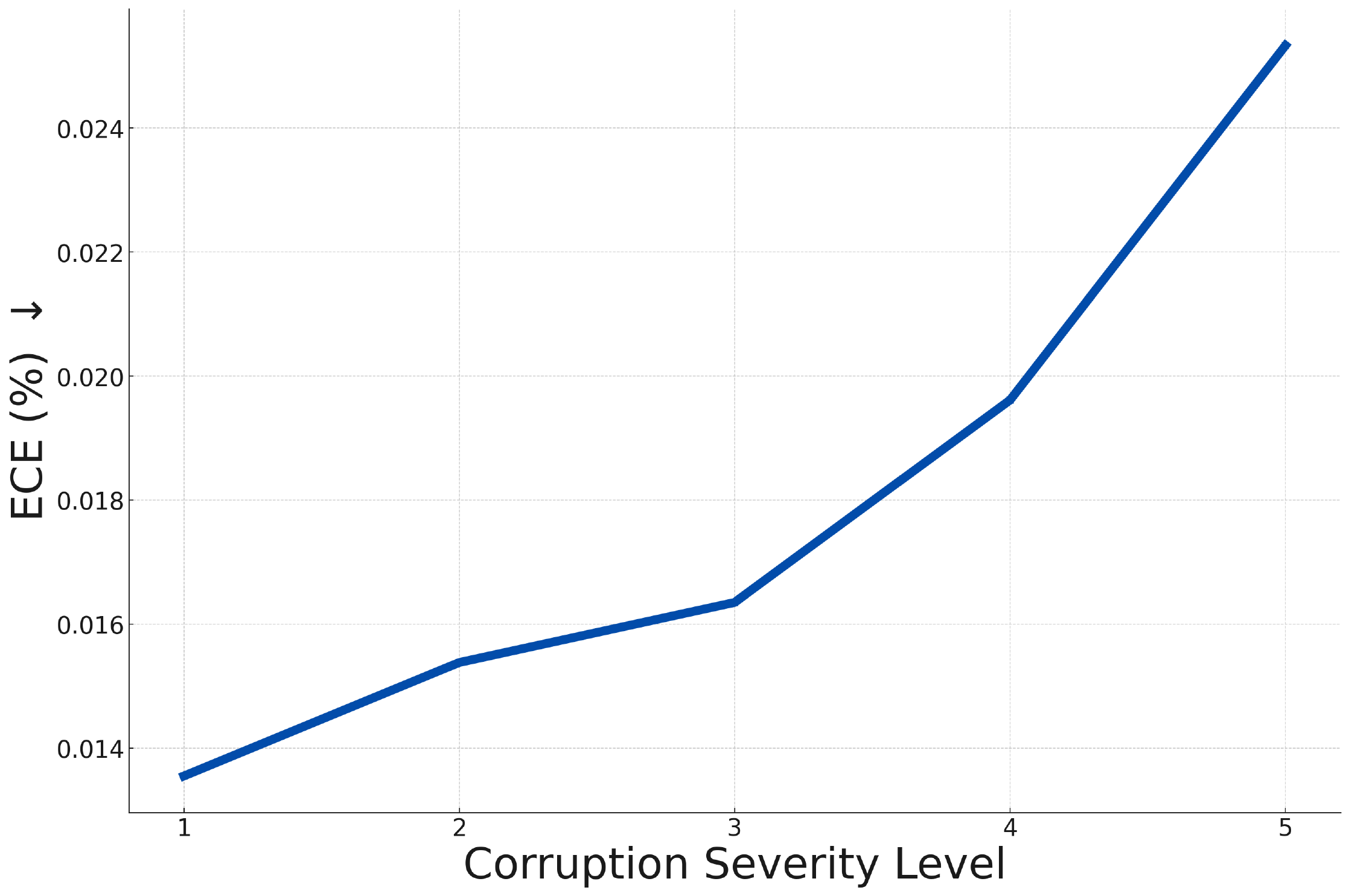} 
        \caption[short caption]{
            \centering
            \begin{minipage}{0.9\linewidth}
                \centering
                Trend of ECE across Dist. shift severity. \\ (Correct Samples)
            \end{minipage}
        }

        \label{fig:shift_correct}
    \end{subfigure}
    \hfill
    \begin{subfigure}[t]{0.32\textwidth} 
        \centering
        \includegraphics[width=\textwidth]{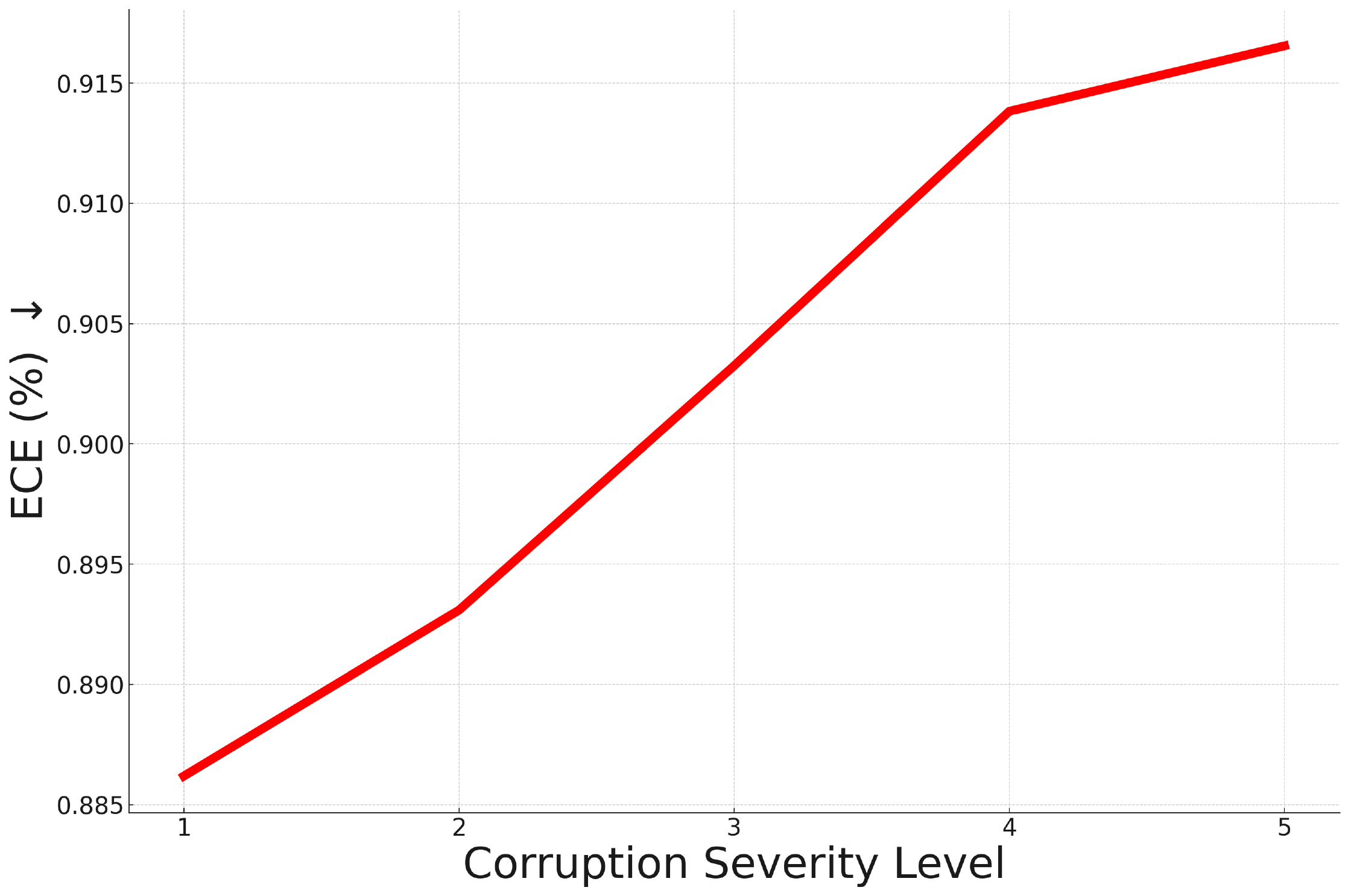} 
        \caption[short caption]{
            \centering
            \begin{minipage}{0.9\linewidth}
                \centering
                Trend of ECE across Dist. shift severity. \\ (Incorrect Samples)
            \end{minipage}
        }
        \label{fig:shift_incorrect}
    \end{subfigure}
    \hfill
    \begin{subfigure}[t]{0.32\textwidth} 
        \centering
        \includegraphics[width=\textwidth]{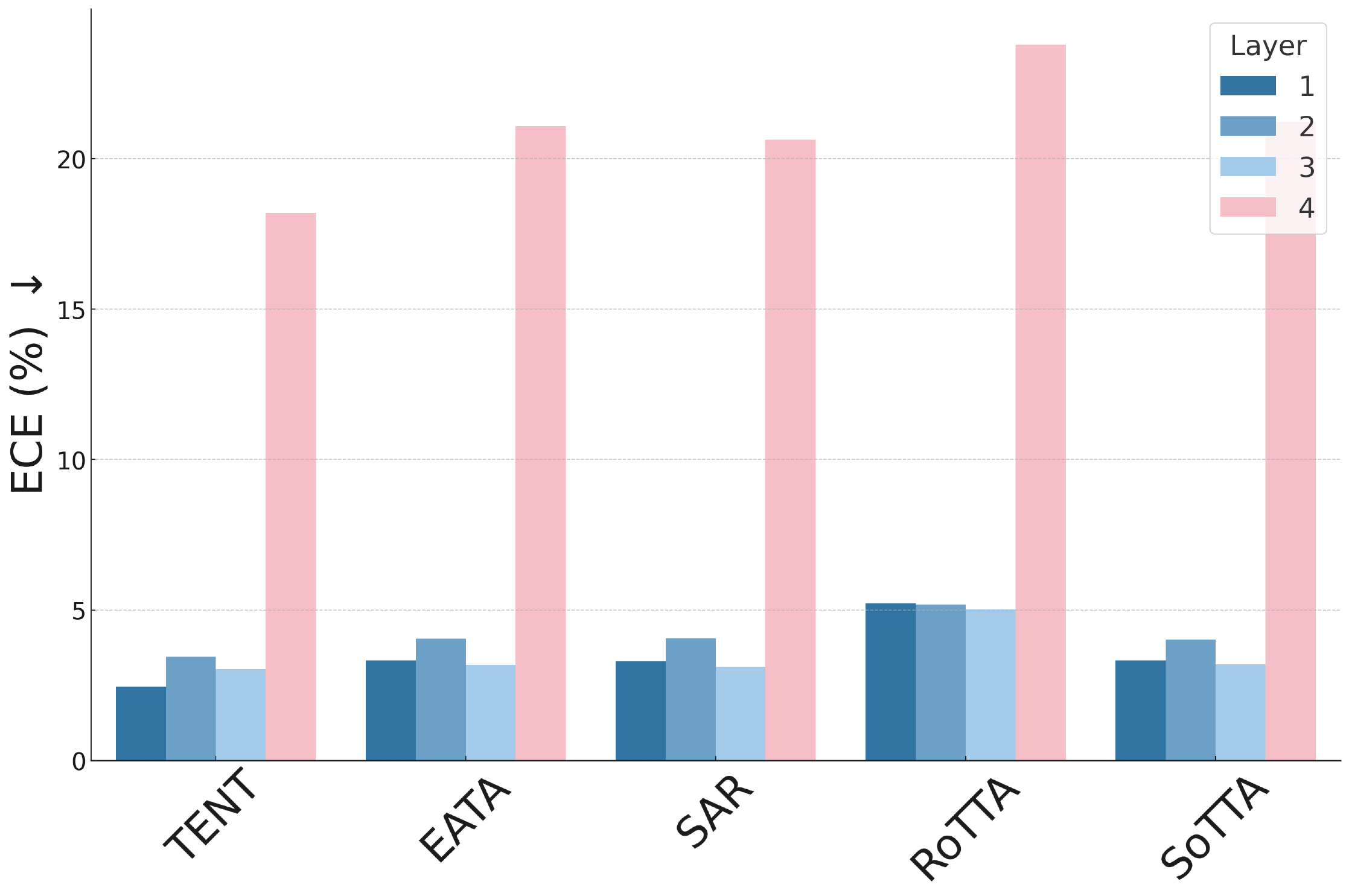} 
        \caption{Ablation for feature extraction layer.}
        \label{fig:ablation_layer}
    \end{subfigure}
    \caption{Comparison of calibration errors across various corruption severity levels for both correct (\ref{fig:shift_correct}) and incorrect (\ref{fig:shift_incorrect}) samples. / Comparison of calibration errors across feature extraction layer for generating style variants of SICL (\ref{fig:ablation_layer})}.
    \label{fig:supp_ablation}
\end{figure}
\definecolor{verylightblue}{RGB}{235, 245, 255} 

\paragraph{Observation - Impact of the distribution shifts on TTA}
We illustrate the impact of distribution shift intensity on calibration performance under TTA environment with representative TTA algorithm TENT~\cite{tent} by plotting the calibration error across corruption severities of all 15 corruptions in the CIFAR10-C dataset in the Figure~\ref{fig:supp_ablation}. Here, we calculated the ECE for correctly predicted samples and incorrectly predicted samples separately, and plotted them in Figure \ref{fig:shift_correct} and Figure \ref{fig:shift_incorrect}, respectively since we wanted to examine the trends of calibration error independently of the model's accuracy.
Regardless of prediction correctness, calibration error rises consistently with increasing shift severity, as models tend to misinterpret novel samples as previously learned representations when shifts deviate further from the training distribution.

\paragraph{Ablation Study - Impact of the feature extraction layer}
We present the ECE for each TTA method in the Figure \ref{fig:ablation_layer}, based on the feature extraction layer used to extract style statistics when creating SICL's Style-shifted Variants. The experiments were conducted on all 15 corruptions of CIFAR-10C, and the average ECE was recorded. Each layer corresponds to a ResNet block. As shown in the figure, extracting features from layers 1, 2, and 3 results in minimal differences in ECE. However, extracting features from layer 4 to create style variants leads to significantly higher errors. This can be attributed to the fact that deeper layers encode more content information~\cite{pan2018two}, rather than style. Consistent with the observations in Section 5.3 of the main paper, this result further demonstrates that when the content of an instance is distorted and used in the ensemble of candidate representations, the method fails to perform proper instance-wise uncertainty estimation.

\section{Further Discussions}
\subsection{Comparison with previous methods}\label{sec:discussion_comparison}
\paragraph{Temperature Scaling-based Methods} 
Temperature scaling-based calibration methods~\cite{guo2017calibration, gal2016dropout, wang2020transferable, hupseudo, park2020calibrated}  assume a fixed model and distribution, scaling the entire target dataset with a single temperature. As a result, they fail to effectively address varying characteristics of each test instances under diverse distribution shifts and changing model during TTA. Moreover, obtaining the temperature often requires additional training with labeled source data~\cite{guo2017calibration, wang2020transferable, park2020calibrated}, or pseudo-data~\cite{hupseudo}, which are unavailable at test time and adds a burden on top of the adaptation process. However, unlike these methods, SICL computes prediction confidence on an instance-by-instance basis, enabling effective uncertainty estimation in highly variable test streams. Since it does not rely on trainable temperatures, it eliminates the need for memory-intensive backward computations or additional trainable parameters.

\paragraph{Ensemble-based Methods}
Ensemble-based calibration methods~\cite{gal2016dropout, lakshminarayanan2017simple} which aggregate stochastic predictions of candidate representations, enable instance-wise calibration, making them capable of handling various non-fixed distributions. However, MC Dropout~\cite{gal2016dropout} can reduce accuracy while averaging logits during dropout inference. And the process of randomly disabling neurons of the model result in representations that lose critical content information (analyzed in Section 5.3). Also, Deep Ensembles~\cite{lakshminarayanan2017simple} require training several models to ensemble, which is undesirable in the resource-restricted test-time adaptation environments. However, unlike these methods, SICL leverages only the forward pass of an online model to compute the consistency ratio across style variants and derives the final prediction confidence. This post-hoc calibration approach does not require additional training or external models and does not compromise accuracy. Furthermore, style-shifted variants of SICL serve as excellent candidate representations because they are generated by extracting feature-level style statistics and perturbing them with bounded Gaussian noise. These representations preserve content information while diversifying only the style.

\subsection{Use case of SICL}
\paragraph{Out-of-distribution Detection}

\begin{figure}[ht!]
    \centering
    \begin{subfigure}[t]{0.45\textwidth} 
        \centering
        \includegraphics[width=\textwidth]{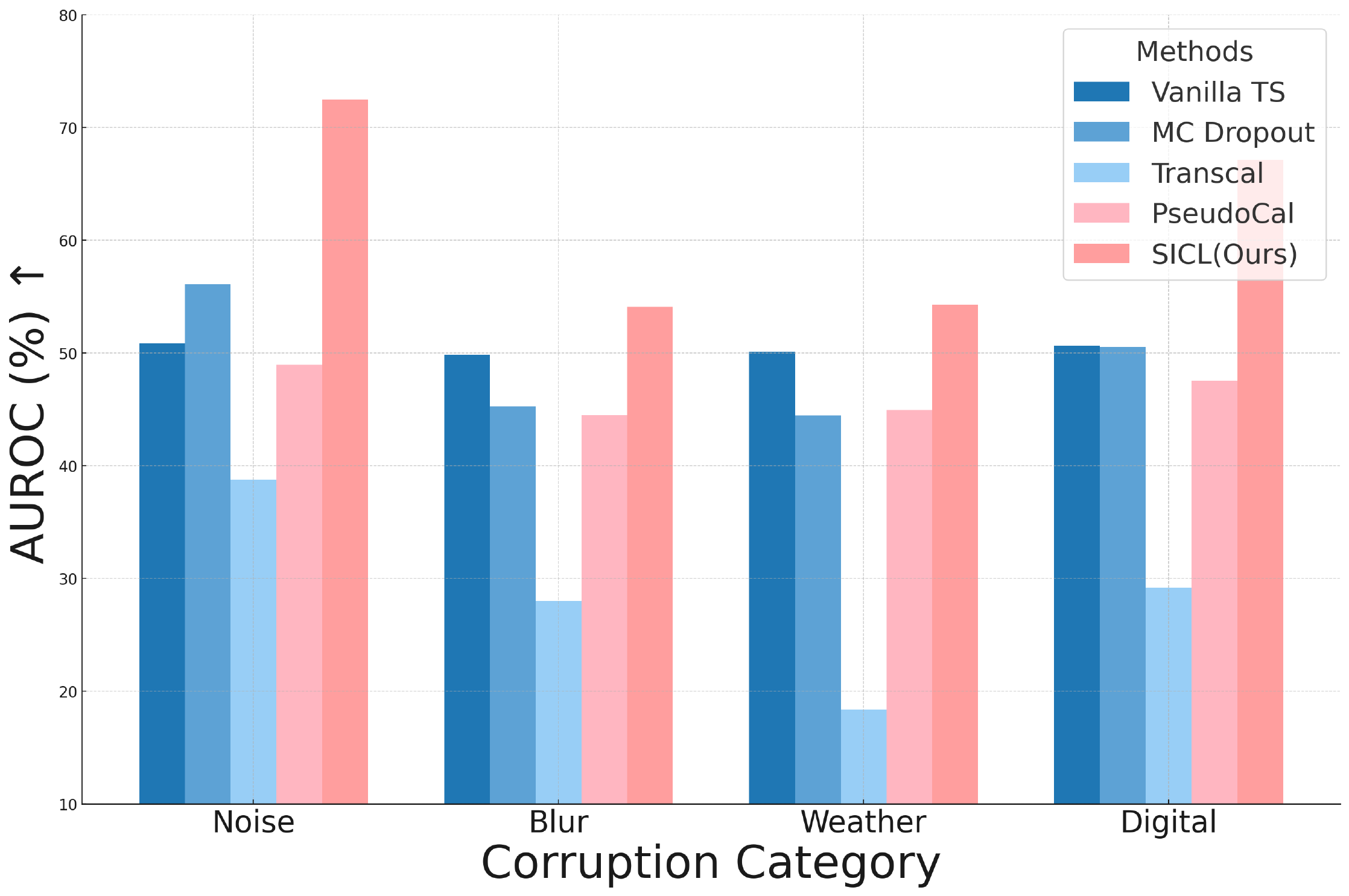} 
        \caption{OOD Detection performance (AUROC(\%))}

        \label{fig:auroc}
    \end{subfigure}
    \hfill
    \begin{subfigure}[t]{0.45\textwidth} 
        \centering
        \includegraphics[width=\textwidth]{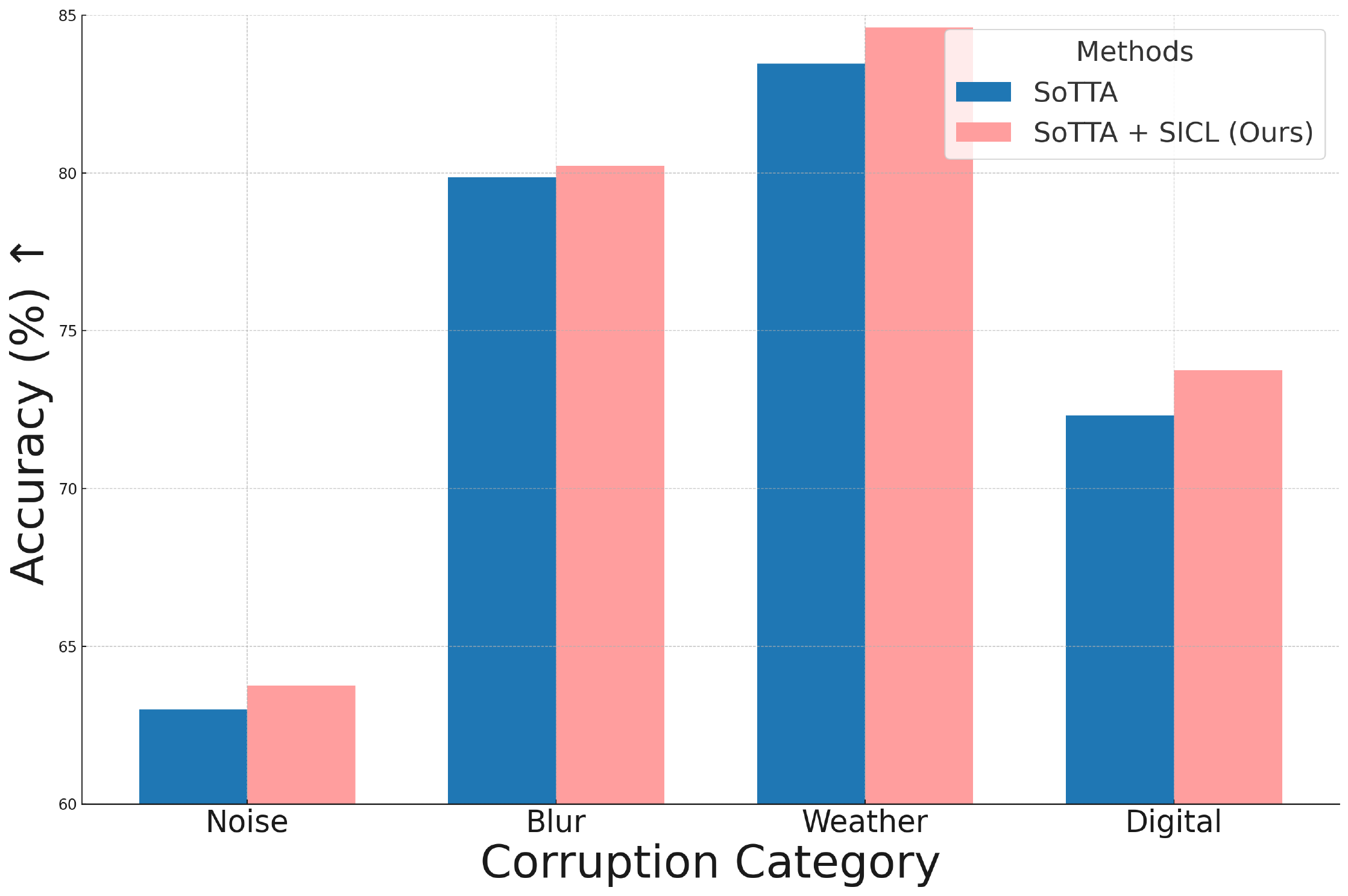} 
        \caption{Impact on classification performance (Accuracy(\%))}
        \label{fig:sottaacc}
    \end{subfigure}
    \caption{Comparison of OOD Detection performance with baselines across various corruption categories (\ref{fig:auroc}) under the noisy stream of CIFAR-10C. / Effect of SICL appilied to SoTTA under the noisy stream of CIFAR-10C (\ref{fig:sottaacc})}.
    \label{fig:supp_d2}
\end{figure}


SICL can naturally be extended to Out-of-Distribution (OOD) detection during TTA. It has been observed in prior study~\cite{sotta}, OOD samples injected during TTA can lead to performance degradation and even model failures. SICL, allowing models to produce reliable predictions, can help mitigate such shortfalls, and give model vendors the freedom of choosing diverse TTA methodologies - instead of relying on a single strategy specifically designed for such scenarios such as SoTTA~\cite{sotta}. In the following paragraphs, we (1) highlight the high-level working of SICL for OOD detection and empirically validate the assertion via (2) OOD detection performance and (3) integration with SoTTA under noisy test streams.

The core concept of SICL lies in estimating correctness likelihood by measuring prediction consistency across style-transformed variants. In-Distribution (ID) samples tend to maintain high consistency than OOD samples, for OOD samples deviate far from the learned content distribution, exhibiting inconsistent predictions under style variants of SICL. Thus, SICL can effectively differentiate between ID and OOD samples via simple thresholding.



We empirically validate this approach by comparing the out-of-distribution detection performance of various calibration methods on the CIFAR-10C \textit{noisy} stream using TENT~\cite{tent}, a representative TTA method, based on AUROC. Figure \ref{fig:auroc} shows that SICL achieved the highest AUROC across all corruption categories. 

We further demonstrate the capability of SICL-based OOD detection method in enhancing performance of TTA methods under noisy test streams in Figure~\ref{fig:sottaacc}. Specifically, we substitute OOD sample filtering process of HUS with SICL - instead of naive filtering of OOD samples based on model's prediction confidence, we apply SICL to calibrate its predictions beforehand. On average, we see accuracy improvements from  75.4\% to 76.4\% in CIFAR10-C corrupted with noisy samples obtained from MNIST. The result demonstrate that through SICL, the model is prevented from learning incorrect representations from noisy samples, thereby enhancing its prediction accuracy.

\paragraph{Uncertainty Modeling in Object Detection}
SICL can potentially be applied to object detection models as well, to determine regions where predictions are uncertain. It is known that object detection models such as YOLO~\cite{redmon2016you} are known to exhibit overconfidence issues~\cite{melotti2022reducing, melotti2023probabilistic}, similar to our observation in TTA methods, hindering its deployment to risk-sensitive domains. Similar to MC Dropblock~\cite{deepshikha2021monte} which utilizes stochastic predictions to model uncertainty in object detection, SICL leverages style perturbations to evaluate prediction consistency, making it possible to identify OOD regions or masks.

\section{Experiment Details}\label{sec:appendix_exp_details}
All our experiments were conducted on NVIDIA TITAN RTX and NVIDIA GeForce RTX 3090 GPUs. 
We use source pre-trained network as an initial model for all datasets.  The source models are trained on clean training data to produce the source models. The training process of the source model employs SGD with a momentum of 0.9 and a cosine annealing learning rate scheduler with the initial learning rate 0.1 over 200 epochs, following SoTTA~\cite{sotta}.



\subsection{Evaluation Details}

To carry out predictive uncertainty calibration in TTA, the model’s predictions are calibrated for each online batch after model updates, ensuring that predictions remain reliable as the model adapts to new distributions. For each test batch at time step \( t \), the predicted confidence \( \text{conf}^{(t)} \) and accuracy \( \text{acc}^{(t)} \) are recorded and accumulated from the beginning to the current time, providing a basis for cumulative calibration error. 
Here, we define Cumulative Expected Calibration Error (ECE) specifically for TTA scenarios, extending the standard ECE definition~\cite{naeini2015obtaining} to quantify the discrepancy between confidence and accuracy over time by aggregating online model's predictions at each time step.
The Cumulative ECE can be expressed as:

\begin{equation}
\begin{aligned}
     \frac{1}{n_T} \! \sum_{t=1}^{n_T} \sum_{k=1}^{K} \frac{|B_k^{(t)}|}{N^{(t)}} \left| \text{acc}_k^{(t)} - \text{conf}_k^{(t)} \right|,
\end{aligned}
\end{equation}
where \( n_T \) represents the total number of test batches,  \( K \) is the number of confidence bins, \( B_k^{(t)} \) is the set of samples in bin \( k \) during the \( t \)-th batch, \(  \text{acc}_k^{(t)}\) and \( \text{conf}_k^{(t)} \) are the accuracy and average confidence of  \( B_k^{(t)} \) and \( N^{(t)} \) is the total number of samples in the \( t \)-th batch. This formulation allows continuous monitoring of calibration quality over time, supporting TTA scenarios where high confidence alignment with accuracy is critical.
Note that all the calibration errors reported in this paper are cumulative ECE.

\subsection{Dataset Details}
\paragraph{CIFAR10-C, CIFAR100-C}
CIFAR10-C and CIFAR100-C are widely recognized benchmark datasets for evaluating model robustness against various types of corruption. Both datasets include 50,000 training samples and 10,000 test samples, divided into 10 and 100 object classes. To test robustness, both datasets introduce 15 corruption types under 4 categories: Noise, Weather, Digital, and Blur. The corruption types are Gaussian Noise, Shot Noise, Impulse Noise, Brightness, Snow, Frost, Fog, Contrast, Elastic Transformation, Pixelation, JPEG Compression, Defocus Blur, Glass Blur, Motion Blur, and Zoom Blur. We use the highest corruption severity level, level 5 for all our experiments, following prior works~\cite{tent, sotta, rotta, schneider2020improving}.

\paragraph{ImageNet-C}
ImageNet-C represents another benchmark widely used to assess model resilience when faced with various corruptions, as noted in several studies [1, 27, 36, 38, 39]. The original ImageNet collection~\cite{deng2009imagenet} encompasses 1,281,167 samples for training and 50,000 for testing purposes. Following a similar approach to CIFAR10-C, this dataset applies an identical set of 15 corruption types, generating a total of 750,000 corrupted test images. In our research, we employ the most severe corruption level (level 5), consistent with our CIFAR10-C and CIFAR100-C setup. 




\subsection{Test Scenario Details}
\begin{figure*}[h]
    \centering
    \includegraphics[width=.8\linewidth]{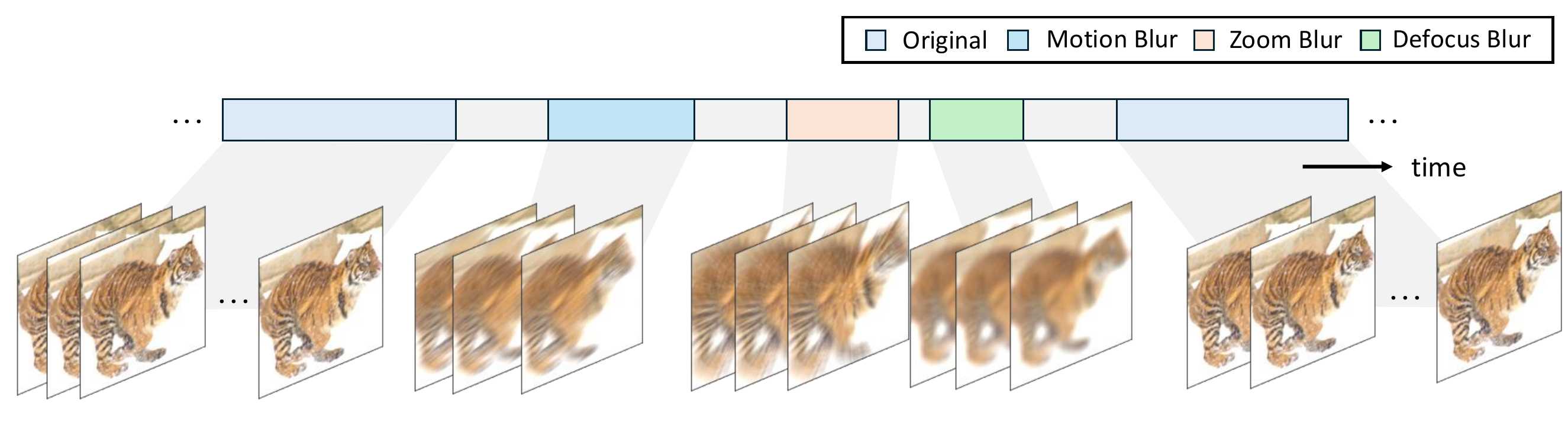}
    \caption{
    An illustration of the \textit{dynamic} test stream over the time axis that could occur when filming moving objects.
    }
    \label{Fig:dynamic_figure}
\end{figure*}
\begin{itemize}
    \item \textbf{Benign}: In the \textit{benign} scenario, we followed the default settings adopted by other TTA works. The test stream was constructed as an i.i.d. distribution from a single corruption type (\textit{e.g.} Gaussian noise) without any noisy samples, and the final results were reported as the average performance across all corruption types. 
    \item \textbf{Dynamic}: To evaluate TTA calibration in realistic scenarios, we introduce a \textbf{dynamic scenario}. Our novel setting shares similarities with prior continual TTA setting, in that it assumes multiple corruption types during test, but differs in the sense that corruption types do not arrive sequentially. Instead, corruption types have temporal correlation generated for each corruption type using a Dirichlet distribution. Figure \ref{Fig:dynamic_figure} demonstrates a visualization of our dynamic test stream scenario.
\end{itemize}

\subsection{Baseline Details}
\paragraph{TTA methods}
Here, we outline the hyperparameter selection of TTA methods. We adopted the hyperparameters documnented in respective papers or source code repositories.
\begin{itemize}
    \item \textbf{TENT.} We set the learning rate as $LR=0.001$ for CIFAR-10-C and CIFAR-100-C, while $LR=0.00025$ for ImageNet-C, adhering to the selection of the original paper. We utilized the original code provided by the authors for implementation.
    \item \textbf{SAR.} We set the learning rate as $LR=(0.00025, 0.001$) for ResNet and ViT models, repectively. Also, we set the sharpness threshold $\rho=0.5$, and entropy threshold $E_0=0.4\times \text{ln}|\mathcal{Y}|$, where $|\mathcal{Y}|$ is the number of classes. We utilized the original code provided by the authors for implementation.
    \item \textbf{EATA.} We set the learning rate as $LR=(0.005, 0.005, 0.00025)$, cosine similaritty threshold $\epsilon=(0.4, 0.4, 0.05)$, tradeoff parameter $\beta=(1, 1, 2000)$ for CIFAR-10-C, CIFAR-100-C and ImageNet-C, respectively. We set the entropy constant $E_0=0.4\times \text{ln}|\mathcal{Y}|$, where $|\mathcal{Y}|$ is the number of classes. We take 2000 samples for calculating Fisher importance, and referred to the official code for implementation. 
    \item \textbf{RoTTA.} We set the learning rate as $LR=0.001$ and $\beta=0.9$, and BN-statistics update moving average update rate $\alpha=0.05$, and Teacher model's exponential moving average updating rate as $\nu=0.001$, and timeliness parameter $\lambda_t=1.0$, and uncertainty parameter $\lambda_u=1.0$, following the authors' selections.
    \item \textbf{SoTTA.} We used a BN momentum of $m = 0.2$, and learning
    rate of $LR = 0.001$ with a single adaptation epoch. We set the HUS size to 64 and the confidence
    threshold $C_0=(0.99, 0.66, 0.33)$ for CIFAR10-C, CIFAR100-C, and ImageNet-C, respectively. We set entropy-sharpness $L_2$-norm constraint $\rho=0.5$ following the suggestion.
\end{itemize}
\paragraph{Calibration methods}
For all calibration baselines and SICL(ours), we performed a hyperparameter search upon representative corruption (Gaussian Noise) and fixed its value across all corruption types.
\begin{itemize}
    \item \textbf{Vanilla TS.} Using the validation set, we optimize the temperature through scikit-learn's optimizer, with initial temperature set as 2.0. 
    \item \textbf{MC Dropout.} We set a fixed dropout ratio of 0.3 and the number of inference steps $N=20$ for all our experiments.
    \item \textbf{TransCal.} As in vanilla TS, we optimize the temperature with the Sequential Least Squares Quadratic Programming (SLSQP) optimization method, adopting the implementation in PseudoCal.
    \item \textbf{PseudoCal.} We set the learning rate of temperature as 0.05, an optimal choice from the CIFAR-10C dataset. We utilized the original code provided by the authors for implementation.
    \item \textbf{SICL(Ours).} We used a fixed number of style variants $N = 20$ and the sensitivity hyperparameter $n=3$ for all our experiments. 
\end{itemize}

\paragraph{Style Shifting methods}
\begin{itemize}
    \item \textbf{MixStyle.} For the qualitative analysis and ablation studies, we set an alpha hyperparameter from beta distribution for deciding mixing ratio as 0.1, adopting the original implementation in MixStyle~\cite{zhou2021domain}.
    
\end{itemize}


\section{Additional Results}\label{sec:appendix_results}
Additionally, to evaluate the performance on a lightweight model, we conducted additional experiments on CIFAR-10C and CIFAR-100C using ResNet-18 as the backbone network. Tables 5 and 6 present the results for CIFAR-10C and CIFAR-100C, respectively. The test stream was evaluated under the Benign setup, while all other settings remained the same.

\begin{table*}[ht]
\centering
\caption{Expected Calibration Error (ECE) (\%) of uncertainty estimation on CIFAR-10C dataset of all corruption types under benign stream using various TTA methods.  \textbf{Bold} numbers represent the lowest error.}
\label{tab:cifar10_ece}
\resizebox{\textwidth}{!}{
\begin{tabular}{cccccccccccccccccc}
\toprule
\multirow{2}{*}{TTA Method} & \multirow{2}{*}{Baseline} & \multicolumn{3}{c}{Noise} & \multicolumn{4}{c}{Blur} & \multicolumn{4}{c}{Weather} & \multicolumn{4}{c}{Digital} & Avg. ($\downarrow$) \\
\cmidrule(lr){3-5} \cmidrule(lr){6-9} \cmidrule(lr){10-13} \cmidrule(lr){14-17}
&  & Gau. & Shot & Imp. & Def. & Gla. & Mot. & Zoom & Snow & Fro. & Fog & Brit. & Cont. & Elas. & Pix. & JPEG & Avg. \\
\midrule
\multirow{5}{*}{TENT~\cite{tent}}
& Vanilla TS & 25.04 & 23.44 & 31.49 & 12.11 & 31.97 & 12.80 & 11.16 & 16.52 & 17.36 & 12.25 & 9.03 & 11.90 & 21.98 & 16.99 & 24.30 & 20.15 \\
& MC Dropout & 15.21 & 14.45 & 19.86 & 7.55 & 22.62 & 7.84 & 6.50 & 10.50 & 11.18 & 7.24 & 5.26 & 8.36 & 13.61 & 10.26 & 16.44 & 11.79 \\
& TransCal & 17.32 & 18.64 & 18.12 & 21.62 & 11.42 & 15.73 & 19.23 & 6.90 & 12.02 & 12.58 & 5.21 & 34.55 & 4.93 & 15.03 & 4.23 & 13.85 \\
& PseudoCal & 8.78 & 7.88 & 12.62 & 3.40 & 14.66 & 3.41 & 3.61 & 4.36 & 4.74 & 3.98 & 3.97 & 4.50 & 6.27 & 3.73 & 8.58 & 6.30 \\
& \cellcolor{verylightblue}SICL(Ours) & \cellcolor{verylightblue}4.78 & \cellcolor{verylightblue}5.23 & \cellcolor{verylightblue}7.55 & \cellcolor{verylightblue}1.20 & \cellcolor{verylightblue}3.01 & \cellcolor{verylightblue}1.52 & \cellcolor{verylightblue}1.51 & \cellcolor{verylightblue}1.30 & \cellcolor{verylightblue}1.41 & \cellcolor{verylightblue}2.17 & \cellcolor{verylightblue}0.98 & \cellcolor{verylightblue}4.40 & \cellcolor{verylightblue}3.52 & \cellcolor{verylightblue}3.05 & \cellcolor{verylightblue}1.69 & \cellcolor{verylightblue} \textbf{2.89} \\
\midrule
\multirow{5}{*}{EATA~\cite{eata}}
& Vanilla TS & 33.13 & 30.41 & 39.24 & 11.68 & 36.09 & 13.42 & 12.53 & 18.60 & 19.73 & 14.74 & 9.16 & 12.87 & 23.18 & 21.43 & 28.06 & 21.55 \\
& MC Dropout & 15.00 & 14.27 & 18.99 & 5.48 & 17.99 & 6.28 & 5.78 & 8.61 & 9.02 & 6.05 & 3.91 & 5.18 & 11.13 & 9.23 & 14.03 & 10.06 \\
& TransCal & 17.12 & 17.09 & 15.60 & 21.95 & 11.22 & 15.12 & 17.78 & 5.55 & 9.53 & 9.97 & 5.15 & 34.12 & 4.01 & 13.67 & 4.04 & 12.86 \\
& PseudoCal & 8.43 & 6.78 & 11.64 & 4.34 & 10.45 & 3.39 & 3.62 & 3.16 & 2.36 & 3.41 & 4.21 & 3.79 & 4.25 & 2.91 & 6.23 & 5.26 \\
& \cellcolor{verylightblue}SICL(Ours) & \cellcolor{verylightblue}7.80 & \cellcolor{verylightblue}8.77 & \cellcolor{verylightblue}6.24 & \cellcolor{verylightblue}1.90 & \cellcolor{verylightblue}3.66 & \cellcolor{verylightblue}1.90 & \cellcolor{verylightblue}1.30 & \cellcolor{verylightblue}1.15 & \cellcolor{verylightblue}3.08 & \cellcolor{verylightblue}2.45 & \cellcolor{verylightblue}1.11 & \cellcolor{verylightblue}2.08 & \cellcolor{verylightblue}4.03 & \cellcolor{verylightblue}4.13 & \cellcolor{verylightblue}3.42 & \cellcolor{verylightblue} \textbf{3.54} \\
\midrule
\multirow{5}{*}{SAR~\cite{sar}}
& Vanilla TS & 31.85 & 30.07 & 35.88 & 11.63 & 33.48 & 13.42 & 12.87 & 18.54 & 19.97 & 14.82 & 9.19 & 12.90 & 23.26 & 21.48 & 28.15 & 21.15 \\
& MC Dropout & 15.30 & 13.97 & 16.88 & 5.46 & 17.82 & 6.21 & 5.77 & 8.66 & 8.93 & 6.15 & 3.88 & 5.30 & 11.24 & 9.27 & 14.26 & 9.94 \\
& TransCal & 17.11 & 17.13 & 16.59 & 21.99 & 11.19 & 15.13 & 17.76 & 5.67 & 9.41 & 9.90 & 5.12 & 34.06 & 4.02 & 13.66 & 4.08 & 13.52 \\
& PseudoCal & 7.79 & 6.69 & 10.73 & 4.21 & 10.19 & 3.20 & 3.68 & 3.36 & 2.59 & 3.38 & 4.20 & 3.64 & 4.10 & 2.75 & 6.32 & 5.12 \\
& \cellcolor{verylightblue}SICL(Ours) & \cellcolor{verylightblue}7.60 & \cellcolor{verylightblue}8.77 & \cellcolor{verylightblue}6.35 & \cellcolor{verylightblue}1.94 & \cellcolor{verylightblue}3.26 & \cellcolor{verylightblue}1.93 & \cellcolor{verylightblue}1.32 & \cellcolor{verylightblue}1.21 & \cellcolor{verylightblue}3.01 & \cellcolor{verylightblue}2.43 & \cellcolor{verylightblue}1.11 & \cellcolor{verylightblue}2.22 & \cellcolor{verylightblue}3.93 & \cellcolor{verylightblue}4.34 & \cellcolor{verylightblue}3.42 & \cellcolor{verylightblue} \textbf{3.52} \\
\midrule
\multirow{5}{*}{RoTTA~\cite{rotta}}
& Vanilla TS & 33.02 & 30.41 & 39.16 & 11.75 & 34.57 & 13.06 & 12.00 & 18.46 & 19.85 & 14.19 & 8.77 & 15.34 & 22.86 & 21.44 & 27.53 & 21.50 \\
& MC Dropout & 13.22 & 12.26 & 17.38 & 5.68 & 16.67 & 5.10 & 4.79 & 7.59 & 7.68 & 4.71 & 2.96 & 10.14 & 9.51 & 9.09 & 13.04 & 9.32 \\
& TransCal & 18.15 & 17.84 & 15.19 & 21.35 & 11.28 & 14.57 & 17.83 & 5.33 & 9.05 & 10.28 & 5.12 & 22.90 & 3.60 & 13.68 & 4.26 & 12.70 \\
& PseudoCal & 8.50 & 7.71 & 12.46 & 4.42 & 10.47 & 3.45 & 4.30 & 3.48 & 3.11 & 3.65 & 4.41 & 4.17 & 4.31 & 3.49 & 6.47 & 5.63 \\
& \cellcolor{verylightblue}SICL(Ours) & \cellcolor{verylightblue}8.88 & \cellcolor{verylightblue}9.20 & \cellcolor{verylightblue}5.79 & \cellcolor{verylightblue}2.39 & \cellcolor{verylightblue}3.57 & \cellcolor{verylightblue}2.82 & \cellcolor{verylightblue}2.14 & \cellcolor{verylightblue}1.49 & \cellcolor{verylightblue}2.65 & \cellcolor{verylightblue}4.21 & \cellcolor{verylightblue}1.79 & \cellcolor{verylightblue}1.35 & \cellcolor{verylightblue}4.24 & \cellcolor{verylightblue}3.91 & \cellcolor{verylightblue}3.57 & \cellcolor{verylightblue} \textbf{4.68} \\
\midrule
\multirow{5}{*}{SoTTA~\cite{sotta}}
& Vanilla TS & 33.46 & 30.95 & 39.32 & 11.63 & 35.08 & 13.42 & 12.87 & 18.54 & 19.97 & 14.82 & 9.19 & 12.90 & 23.26 & 21.48 & 28.15 & 21.66 \\
& MC Dropout & 15.84 & 14.44 & 19.48 & 5.30 & 19.08 & 6.32 & 9.27 & 5.89 & 9.26 & 7.04 & 4.03 & 5.45 & 11.65 & 10.67 & 14.81 & 10.57 \\
& TransCal & 17.05 & 17.18 & 15.43 & 21.99 & 11.23 & 15.13 & 17.76 & 5.67 & 9.41 & 9.90 & 5.12 & 34.06 & 4.02 & 13.66 & 4.09 & 13.44 \\
& PseudoCal & 8.91 & 11.94 & 6.86 & 4.21 & 10.38 & 3.28 & 3.61 & 3.32 & 2.68 & 3.39 & 4.01 & 3.64 & 4.31 & 2.82 & 6.37 & 5.35 \\
& \cellcolor{verylightblue}SICL(Ours) & \cellcolor{verylightblue}8.20 & \cellcolor{verylightblue}6.16 & \cellcolor{verylightblue}8.77 & \cellcolor{verylightblue}1.94 & \cellcolor{verylightblue}3.67 & \cellcolor{verylightblue}1.93 & \cellcolor{verylightblue}1.32 & \cellcolor{verylightblue}1.21 & \cellcolor{verylightblue}3.01 & \cellcolor{verylightblue}2.43 & \cellcolor{verylightblue}1.11 & \cellcolor{verylightblue}2.22 & \cellcolor{verylightblue}3.93 & \cellcolor{verylightblue}4.34 & \cellcolor{verylightblue}3.42 & \cellcolor{verylightblue} \textbf{3.58} \\
\bottomrule
\end{tabular}
}
\end{table*}

\begin{table*}[ht]
\centering
\caption{Expected Calibration Error (ECE) (\%) of uncertainty estimation on CIFAR-100C dataset of all corruption types under benign stream using various TTA methods.  \textbf{Bold} numbers represent the lowest error.}
\label{tab:cifar100_ece}
\resizebox{\textwidth}{!}{
\begin{tabular}{ccccccccccccccccccc}
\toprule
\multirow{2}{*}{TTA Method} & \multirow{2}{*}{Baseline} & \multicolumn{3}{c}{Noise} & \multicolumn{4}{c}{Blur} & \multicolumn{4}{c}{Weather} & \multicolumn{4}{c}{Digital} & Avg. ($\downarrow$) \\
\cmidrule(lr){3-5} \cmidrule(lr){6-9} \cmidrule(lr){10-13} \cmidrule(lr){14-17}
 &  & Gau. & Shot & Imp. & Def. & Gla. & Mot. & Zoom & Snow & Fro. & Fog & Brit. & Cont. & Elas. & Pix. & JPEG & Avg. \\
\midrule
\multirow{5}{*}{TENT~\cite{tent}}
& Vanilla TS & 49.10 & 33.16 & 38.83 & 21.82 & 35.29 & 23.21 & 20.77 & 27.41 & 27.68 & 24.84 & 19.84 & 23.03 & 28.07 & 24.58 & 30.82 & 28.56 \\
& MC Dropout & 5.85 & 6.20 & 4.31 & 5.48 & 6.91 & 6.08 & 6.11 & 6.06 & 5.14 & 5.26 & 7.26 & 6.30 & 5.33 & 7.05 & 4.76 & 5.79 \\
& TransCal & 34.10 & 35.68 & 32.95 & 49.91 & 34.78 & 44.14 & 49.99 & 39.13 & 43.38 & 42.19 & 36.81 & 48.71 & 34.12 & 48.83 & 34.72 & 40.63 \\
& PseudoCal & 8.78 & 8.76 & 7.80 & 12.83 & 8.93 & 11.80 & 13.77 & 11.12 & 11.49 & 12.17 & 13.12 & 11.92 & 11.24 & 12.85 & 10.16 & 11.12 \\
& \cellcolor{verylightblue}SICL(Ours) & \cellcolor{verylightblue}6.65 & \cellcolor{verylightblue}6.88 & \cellcolor{verylightblue}7.12 & \cellcolor{verylightblue}2.84 & \cellcolor{verylightblue}2.50 & \cellcolor{verylightblue}5.22 & \cellcolor{verylightblue}7.07 & \cellcolor{verylightblue}7.04 & \cellcolor{verylightblue}5.20 & \cellcolor{verylightblue}3.31 & \cellcolor{verylightblue}6.27 & \cellcolor{verylightblue}4.86 & \cellcolor{verylightblue}8.34 & \cellcolor{verylightblue}7.36 & \cellcolor{verylightblue}2.31 & \cellcolor{verylightblue} \textbf{5.49} \\
\midrule
\multirow{5}{*}{EATA~\cite{eata}}
& Vanilla TS & 52.76 & 50.97 & 55.07 & 31.66 & 53.62 & 38.65 & 32.06 & 43.69 & 38.88 & 37.74 & 29.38 & 30.65 & 46.45 & 40.09 & 48.30 & 41.99 \\
& MC Dropout & 15.49 & 16.45 & 14.46 & 17.08 & 17.37 & 20.27 & 15.91 & 19.56 & 17.05 & 18.38 & 25.60 & 16.96 & 16.53 & 15.85 & 17.17 & 17.64 \\
& TransCal & 18.91 & 21.93 & 12.13 & 34.65 & 11.87 & 28.97 & 29.74 & 16.51 & 27.08 & 26.56 & 21.98 & 36.11 & 11.80 & 24.66 & 13.14 & 22.40 \\
& PseudoCal & 10.74 & 10.22 & 10.26 & 9.69 & 13.61 & 9.61 & 9.47 & 9.42 & 8.70 & 8.89 & 9.74 & 8.73 & 9.87 & 8.88 & 9.64 & 9.83 \\
& \cellcolor{verylightblue}SICL(Ours) & \cellcolor{verylightblue}5.83 & \cellcolor{verylightblue}7.55 & \cellcolor{verylightblue}12.98 & \cellcolor{verylightblue}2.80 & \cellcolor{verylightblue}17.20 & \cellcolor{verylightblue}3.79 & \cellcolor{verylightblue}4.82 & \cellcolor{verylightblue}5.19 & \cellcolor{verylightblue}8.84 & \cellcolor{verylightblue}5.35 & \cellcolor{verylightblue}3.84 & \cellcolor{verylightblue}3.79 & \cellcolor{verylightblue}11.19 & \cellcolor{verylightblue}4.12 & \cellcolor{verylightblue}12.95 & \cellcolor{verylightblue} \textbf{7.61} \\
\midrule
\multirow{5}{*}{SAR~\cite{sar}}
& Vanilla TS & 33.74 & 32.19 & 37.25 & 22.01 & 35.69 & 24.16 & 21.87 & 28.07 & 28.59 & 25.84 & 20.05 & 20.50 & 29.07 & 24.77 & 31.38 & 27.68 \\
& MC Dropout & 8.62 & 9.40 & 7.40 & 10.59 & 11.29 & 11.87 & 8.55 & 11.14 & 9.93 & 11.87 & 11.48 & 24.36 & 11.15 & 10.39 & 10.89 & 11.24 \\
& TransCal & 34.93 & 36.84 & 35.36 & 51.23 & 36.73 & 44.58 & 49.88 & 39.86 & 43.05 & 42.57 & 37.82 & 51.26 & 34.71 & 49.84 & 35.86 & 41.57 \\
& PseudoCal & 7.94 & 7.62 & 7.48 & 9.98 & 7.19 & 8.49 & 10.04 & 9.35 & 8.49 & 9.48 & 10.25 & 10.43 & 9.60 & 9.59 & 8.36 & 8.95 \\
& \cellcolor{verylightblue}SICL(Ours) & \cellcolor{verylightblue}7.49 & \cellcolor{verylightblue}8.07 & \cellcolor{verylightblue}6.22 & \cellcolor{verylightblue}6.76 & \cellcolor{verylightblue}2.74 & \cellcolor{verylightblue}3.15 & \cellcolor{verylightblue}4.51 & \cellcolor{verylightblue}2.07 & \cellcolor{verylightblue}4.98 & \cellcolor{verylightblue}6.98 & \cellcolor{verylightblue}4.28 & \cellcolor{verylightblue}6.11 & \cellcolor{verylightblue}8.69 & \cellcolor{verylightblue}8.06 &\cellcolor{verylightblue}2.59 & \cellcolor{verylightblue} \textbf{5.64} \\
\midrule
\multirow{5}{*}{RoTTA~\cite{rotta}}
& Vanilla TS & 39.81 & 38.98 & 42.62 & 22.64 & 38.83 & 24.21 & 22.91 & 30.58 & 30.94 & 27.92 & 20.83 & 23.60 & 30.03 & 28.49 & 35.62 & 30.53 \\
& MC Dropout & 11.14 & 12.28 & 9.32 & 7.65 & 9.28 & 6.91 & 9.10 & 7.64 & 7.36 & 11.03 & 7.71 & 9.92 & 10.82 & 10.88 & 7.86 & 8.59 \\
& TransCal & 25.95 & 25.82 & 22.19 & 46.74 & 27.02 & 40.66 & 45.13 & 31.15 & 29.04 & 36.70 & 32.89 & 17.73 & 29.29 & 39.19 & 24.37 & 31.62 \\
& PseudoCal & 7.98 & 8.64 & 8.04 & 9.71 & 8.42 & 9.29 & 9.86 & 10.15 & 9.01 & 10.46 & 10.39 & 9.25 & 9.83 & 10.08 & 8.47 & 9.30 \\
& \cellcolor{verylightblue}SICL(Ours) & \cellcolor{verylightblue}9.07 & \cellcolor{verylightblue}10.08 & \cellcolor{verylightblue}6.84 & \cellcolor{verylightblue}7.70 & \cellcolor{verylightblue}6.05 & \cellcolor{verylightblue}6.46 & \cellcolor{verylightblue}5.95 & \cellcolor{verylightblue}5.35 & \cellcolor{verylightblue}5.31 & \cellcolor{verylightblue}8.89 & \cellcolor{verylightblue}4.16 & \cellcolor{verylightblue}3.78 & \cellcolor{verylightblue}9.05 & \cellcolor{verylightblue}10.52 & \cellcolor{verylightblue}3.10 & \cellcolor{verylightblue} \textbf{8.37} \\
\midrule
\multirow{5}{*}{SoTTA~\cite{sotta}}
& Vanilla TS & 40.92 & 39.41 & 43.62 & 23.11 & 39.04 & 25.67 & 24.05 & 31.60 & 32.20 & 28.78 & 21.65 & 23.69 & 31.68 & 29.13 & 36.69 & 31.25 \\
& MC Dropout & 8.73 & 9.87 & 8.57 & 5.77 & 6.40 & 6.49 & 5.67 & 6.71 & 5.82 & 6.08 & 5.67 & 9.46 & 6.93 & 7.08 & 6.04 & 6.57 \\
& TransCal & 27.28 & 28.11 & 20.47 & 48.62 & 29.20 & 40.79 & 46.41 & 33.64 & 37.19 & 36.68 & 35.43 & 45.72 & 30.28 & 42.49 & 25.45 & 35.50 \\
& PseudoCal & 7.44 & 10.58 & 7.99 & 9.07 & 9.79 & 9.12 & 9.05 & 9.46 & 9.37 & 9.08 & 10.06 & 9.46 & 9.37 & 9.80 & 9.25 & 9.03 \\
& \cellcolor{verylightblue}SICL(Ours) & \cellcolor{verylightblue}7.65 & \cellcolor{verylightblue}9.07 & \cellcolor{verylightblue}5.97 & \cellcolor{verylightblue}5.77 & \cellcolor{verylightblue}4.89 & \cellcolor{verylightblue}4.53 & \cellcolor{verylightblue}5.35 & \cellcolor{verylightblue}5.08 & \cellcolor{verylightblue}6.13 & \cellcolor{verylightblue}7.18 & \cellcolor{verylightblue}4.95 & \cellcolor{verylightblue}4.10 & \cellcolor{verylightblue}7.71 & \cellcolor{verylightblue}9.25 & \cellcolor{verylightblue}3.37 & \cellcolor{verylightblue} \textbf{6.19} \\

\bottomrule
\end{tabular}
}
\end{table*}


\section{License of Assets}
\subsection{Datasets}
CIFAR10/CIFAR100 (MIT License), CIFAR10-C/CIFAR100-C (Creative Commons
Attribution 4.0 International), MNIST (CC-BY-NC-SA 3.0), ImageNet-C (Apache 2.0).
\subsection{Codes}
Torchvision for ResNet-18, ResNet-50, and ResNet-101 (Apache 2.0), the vit-pytorch repository from lucidrains (MIT License), the official repository of CoTTA (MIT License), the official repository of TENT (MIT License), the official repository of EATA (MIT License), the official repository of SAR (BSD 3-Clause License), the official repository of RoTTA (MIT License), the official repository of SoTTA (MIT License), and the official repository of PseudoCal (MIT License).

\end{document}